%% file: main.tex
\documentclass[10pt,journal,compsoc]{IEEEtran}

\ifCLASSOPTIONcompsoc
  \usepackage[nocompress]{cite}
\else
  \usepackage{cite}
\fi
\ifCLASSINFOpdf
  \usepackage[pdftex]{graphicx}
  \DeclareGraphicsExtensions{.pdf,.jpeg,.png}
\else
  \usepackage[dvips]{graphicx}
  \DeclareGraphicsExtensions{.eps}
\fi
\usepackage{amsmath,amssymb,amsfonts,dsfont,pifont,bm,bbm,mathrsfs,mathtools}
\usepackage{algorithm,algpseudocode,listings}
\usepackage{booktabs,multirow,adjustbox,diagbox}
\usepackage{xspace}
\usepackage{float,capt-of}
\usepackage[table]{xcolor}
\usepackage{url}
\usepackage{makecell}
\usepackage[pagebackref=false,breaklinks=true,colorlinks=true,citecolor=blue,bookmarks=false]{hyperref}
\usepackage[capitalize]{cleveref}  %
\usepackage{tabularray}
\usepackage{subcaption}
\crefname{section}{Sec.}{Secs.}
\Crefname{section}{Section}{Sections}
\crefname{table}{Tab.}{Tabs.}
\Crefname{table}{Table}{Tables}
\crefname{figure}{Fig.}{Figs.}
\Crefname{figure}{Figure}{Figures}
\crefname{equation}{Eq.}{Eqs.}
\Crefname{equation}{Equation}{Equations}
\newcommand{\z}{{\rm\bf z}}         %
\newcommand{\x}{{\rm\bf x}}         %
\newcommand{\Loss}{\mathcal{L}}     %
\newcommand{\E}{\mathbb{E}}         %

\newcommand{\etal}{\textit{et al.}}

\newcommand{\tocite}[1]{\textcolor{red}{[TOCITE]}}

\usepackage{tikz}
\usetikzlibrary{tikzmark, calc}
\tikzset{
    double color fill/.code 2 args={
        \pgfdeclareverticalshading[%
            tikz@axis@top,tikz@axis@middle,tikz@axis@bottom%
        ]{diagonalfill}{100bp}{%
            color(0bp)=(tikz@axis@bottom);
            color(50bp)=(tikz@axis@bottom);
            color(50bp)=(tikz@axis@middle);
            color(50bp)=(tikz@axis@top);
            color(100bp)=(tikz@axis@top)
        }
        \tikzset{shade, left color=#1, right color=#2, shading=diagonalfill}
    }
}
\tikzset{%
    diagonal fill/.style 2 args={%
        double color fill={#1}{#2},
        shading angle=45,
        opacity=0.8},
    filling 1/.style={%
        shade,
        shading=myshade1,
        shading angle=0,
        opacity=0.5},
    filling 2/.style={%
        shade,
        shading=myshade2,
        shading angle=0,
        opacity=0.5},
    filling 3/.style={%
        shade,
        shading=myshade3,
        shading angle=0,
        opacity=0.5}
   }
\pgfdeclarehorizontalshading{myshade1}{100bp}{%
      color(0bp)=(orange!30);
      color(50bp)=(orange!30);
      color(50bp)=(yellow!30);
      color(100bp)=(yellow!30)}

\pgfdeclarehorizontalshading{myshade2}{100bp}{%
      color(0bp)=(orange!30);
      color(50bp)=(orange!30);
      color(50bp)=(red!30);
      color(100bp)=(red!30)}
\pgfdeclarehorizontalshading{myshade3}{100bp}{%
      color(0bp)=(orange!30);
      color(50bp)=(orange!30);
      color(50bp)=(cyan!30);
      color(100bp)=(cyan!30)}
\newcounter{BGnum}
\definecolor{color1}{rgb}{0.87, 0.894, 0.768}
\definecolor{color2}{rgb}{0.90,0.773,0.773}%
\definecolor{color3}{rgb}{0.773, 0.824, 0.886}
\definecolor{color4}{rgb}{0.95, 0.83, 0.71}

\usepackage{dcolumn, array, booktabs}
\newcolumntype{.}{D{.}{.}{-1}}

\newcommand{\com}[1]{\colorbox{color1}{#1}}
\newcommand{\glo}[1]{\colorbox{white}{#1}}
\newcommand{\gloreli}[1]{\colorbox{color2}{#1}}%
\newcommand{\glohuman}[1]{\colorbox{color3}{#1}}%
\newcommand{\gloseman}[1]{\colorbox{color4}{#1}}%

\begin{document}

\newcommand{\titlename}{3D Generative Models: A Survey}
\title{\titlename}

\author{
Zifan Shi*,
Sida Peng*,
Yinghao Xu*, 
Andreas Geiger,
Yiyi Liao, and
Yujun Shen
\IEEEcompsocitemizethanks{
    \IEEEcompsocthanksitem Z. Shi is with
        the Department of Computer Science and Engineering, the Hong Kong University of Science and Technology, Hong Kong SAR. \protect
    \IEEEcompsocthanksitem S. Peng is with
        School of Software Technology, Zhejiang University, China.\protect
    \IEEEcompsocthanksitem Y. Xu is with
        the Department of Information Engineering, the Chinese University of Hong Kong, Hong Kong SAR. \protect
    \IEEEcompsocthanksitem A. Geiger is with
        the  Autonomous Vision Group, University of Tubingen and Tubingen AI Center, Germany.\protect
    \IEEEcompsocthanksitem Y. Liao is with
        the College of Information Science \& Electronic Engineering, Zhejiang University, China. She is the co-corresponding author:  \href{mailto:yiyi.liao@zju.edu.cn}{yiyi.liao@zju.edu.cn} \protect
    \IEEEcompsocthanksitem Y. Shen is with
        Ant Group, China. He is the corresponding author: \href{mailto:shenyujun0302@gmail.com}{shenyujun0302@gmail.com} \protect
  }%
  \thanks{* denotes equal contribution.}  %
}

\IEEEtitleabstractindextext{
    \input{sections/00_abstract.tex}
}

\maketitle
\IEEEdisplaynontitleabstractindextext
\IEEEpeerreviewmaketitle

\input{sections/01_introduction.tex}
\input{sections/02_scope.tex}
\input{sections/03_fundamentals.tex}
\input{sections/04_shape_generation.tex}

\input{sections/05_image_generation.tex}

\input{sections/06_applications.tex}

\input{sections/07_future_work.tex}

\input{sections/08_conclusion.tex}

\ifCLASSOPTIONcaptionsoff
  \newpage
\fi
\bibliographystyle{IEEEtran}
\bibliography{references}

\end{document}

%% file: sections/00_abstract.tex
\parbox{0.918\textwidth}{
\begin{abstract}
Generative models aim to learn the distribution of observed data by generating new instances. 
With the advent of neural networks, deep generative models, including variational autoencoders (VAEs), generative adversarial networks (GANs), and diffusion models (DMs), have progressed remarkably in synthesizing 2D images. Recently, researchers started to shift focus from 2D to 3D space, considering that 3D data is more closely aligned with our physical world and holds immense practical potential. However, unlike 2D images, which possess an inherent and efficient representation (\textit{i.e.}, a pixel grid), representing 3D data poses significantly greater challenges. Ideally, a robust 3D representation should be capable of accurately modeling complex shapes and appearances while being highly efficient in handling high-resolution data with high processing speeds and low memory requirements. Regrettably, existing 3D representations, such as point clouds, meshes, and neural fields, often fail to satisfy all of these requirements simultaneously. In this survey, we thoroughly review the ongoing developments of 3D generative models, including methods that employ 2D and 3D supervision. Our analysis centers on generative models, with a particular focus on the representations utilized in this context. We believe our survey will help the community to track the field's evolution and to spark innovative ideas to propel progress towards solving this challenging task.

\end{abstract}
}

%% file: sections/01_introduction.tex
\input{figures/general_pipeline.tex}

\IEEEraisesectionheading{\section{Introduction}\label{sec:introduction}}

\IEEEPARstart{T}{he} rapid advancement of deep learning~\cite{lecun2015deep} has revolutionized various computer vision tasks, such as visual object recognition~\cite{alexnet,vggnet}, object detection~\cite{fastrcnn,fasterrcnn,maskrcnn}, and image rendering~\cite{unisurf,volsdf,nerf}. It has also brought significant improvements to our daily lives, enabling autonomous driving~\cite{pointfusion,argoverse}, advancements in biological research~\cite{alphafold}, and facilitating intelligent creations~\cite{dalle,dalle2}.

Among the different techniques, generative modeling~\cite{gan,vae,DDPM} holds a crucial position in data analysis and machine learning. Unlike discriminative models that focus on direct predictions, generative models aim to capture the underlying data distribution to generate new instances. Consequently, they require a comprehensive understanding of the data.
For instance, while a recognition model might disregard task-irrelevant details (\textit{e.g.}, color) or suffer from shortcut learning \cite{geirhos2020shortcut}, a generative model is expected to model every intricate detail (\textit{e.g.}, object arrangement and texture) to achieve convincing results. From this perspective, learning a generative model is usually more challenging yet facilitates a range of applications~\cite{ghfeat,pix2pix,interfacegan,dalle2} and is important for generalization and robustness.

\begin{figure*}[t]
    \centering
    \includegraphics[width=1.0\linewidth]{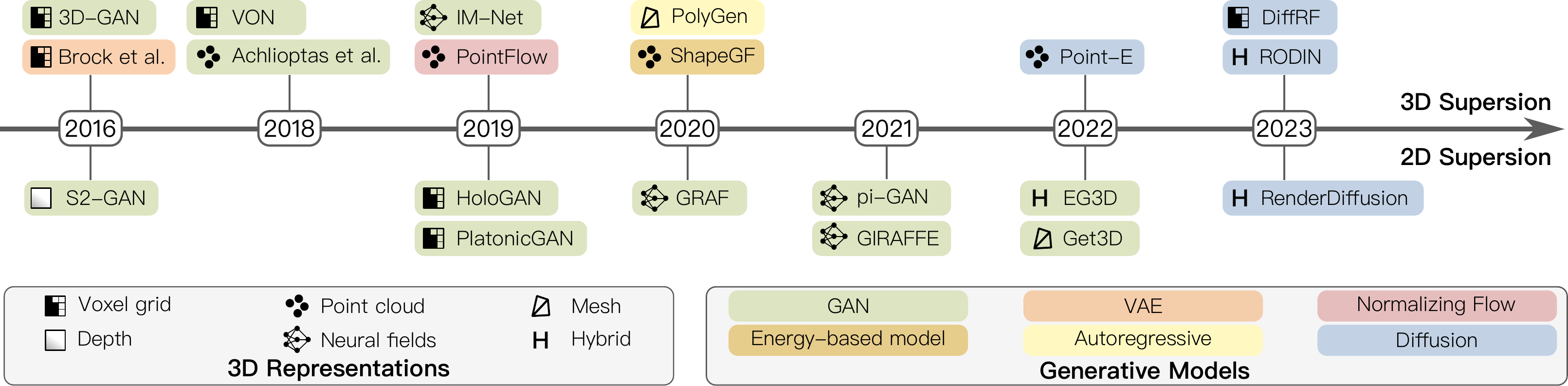}
    \caption{
       \textbf{3D generative model timeline.} We show representative methods trained with 3D supervision (top) and 2D supervision (bottom), respectively. Each method is illustrated with its 3D representation and the generative model.
    }
    \label{fig:generative-models} 
\end{figure*}

The past few years have witnessed the incredible success of deep generative models~\cite{vae,gan,DDPM} in 2D image synthesis~\cite{dalle2, imagen, parti}.
Despite their different formulations, variational autoencoders (VAEs)~\cite{vae}, autoregressive models (ARs)~\cite{AR}, normalizing flows (NFs)~\cite{normalizingflows}, generative adversarial networks (GANs)~\cite{gan}, and the very recent diffusion probabilistic models (DPMs)~\cite{DDPM} all possess the ability to transform latent variables into high-quality images.
Nowadays, however, the application of generative models exclusively in 2D falls short in meeting the demands of several important real-world scenarios which require access to 3D information relevant to modeling the physical image formation process. Taking the film industry as a prime example, the need arises to design 3D digital assets instead of merely producing 2D images, enabling more immersive experiences.
Existing content creation pipelines often rely on significant expertise and extensive human effort, resulting in time-consuming and costly processes. Numerous pioneering attempts~\cite{wu2016learning, achlioptas2018learning, chen2019learning, graf, pigan, giraffe}, as shown in \cref{fig:generative-models}, have been made to explore automated 3D data generation. However, these studies are still in their early stages, with much progress yet to be made.

One of the key distinctions between 2D generation and 3D generation lies in the data format. Specifically, a 2D image can be naturally represented as an array of pixel values, which can be conveniently processed by neural networks~\cite{alexnet,vggnet}. On the other hand, there are numerous 3D representations available to depict a 3D instance, such as point clouds~\cite{pointnet,pointnet++}, meshes~\cite{surfnet,pixel2mesh}, voxel grids~\cite{3dshapenet,voxnet}, multi-plane images~\cite{MPI}, implicit neural representations~\cite{nerf}, and so on. Each representation possesses its own advantages and limitations. For instance, meshes offer a compact representation of 3D shapes but are challenging to analyze and generate using neural networks due to their irregular data structure. In contrast, voxel grids are regularly aligned in 3D space and work well with standard 3D convolutional neural networks. However, they are memory-consuming and cannot represent high-resolution 3D scenes without sophisticated sparse data structures. Hence, the selection of an appropriate representation is crucial in 3D content generation.

Considering the rapidly growing field of 3D generative models, this paper offers a comprehensive survey to educate beginners and serve as a reference for practitioners and experts alike. While there already exist several surveys in the literature investigating generative models~\cite{deepgenerativemodelssurvey,generativemodellingsurvey}, 3D vision~\cite{neuralrenderingsurvey,pointcloudsurvey,3dvisiontransformersurvey, xie2022neural}, as well as generation of 3D structures~\cite{Chaudhuri2020CGF} and faces~\cite{toshpulatov2021generative}, a comprehensive review of 3D generative models is still missing.
As discussed earlier, accomplishing such a challenging task involves considering numerous candidate generative models (\textit{e.g.}, VAEs, and GANs) and representations (\textit{e.g.}, point clouds, and implicit neural representations).
To overcome this problem, our survey categorizes 3D generative models according to 3D representations and summarizes the general pipelines of these methods, as presented in \cref{fig:general_pipeline}, \cref{fig:3d_generator}, and \cref{fig:3dgan_pipeline}.
Our survey clearly illustrates the suitability of specific 3D representations for various types of generative models, taking into account factors such as neural network compatibility, memory efficiency, and representation capability.
In addition, we classify generative models based on the supervision signal and illustrate design choices of generative models with or without 3D supervision.
Furthermore, we compare different 3D generative models quantitatively in terms of generation capability and efficiency. 
We believe our survey helps the reader to grasp a more comprehensive understanding of the field of 3D generative models.

The remainder of this paper is organized as follows: \cref{sec:scope} clarifies the scope of this survey. \cref{sec:fundamentals} introduces the fundamentals of the 3D generation task, including formulations of various generative models and popular 3D representations. \cref{sec:shape_generation} and \cref{sec:image_generation} summarize existing approaches for learning 3D representations from 2D and 3D data, respectively. \cref{sec:applications} discusses the downstream applications of 3D generative models. Finally, \cref{sec:future_work} discusses unsolved problems and future work in the field of 3D generative models.

%% file: figures/general_pipeline.tex
\begin{figure*}[t]
    \centering
    \includegraphics[width=1\textwidth]{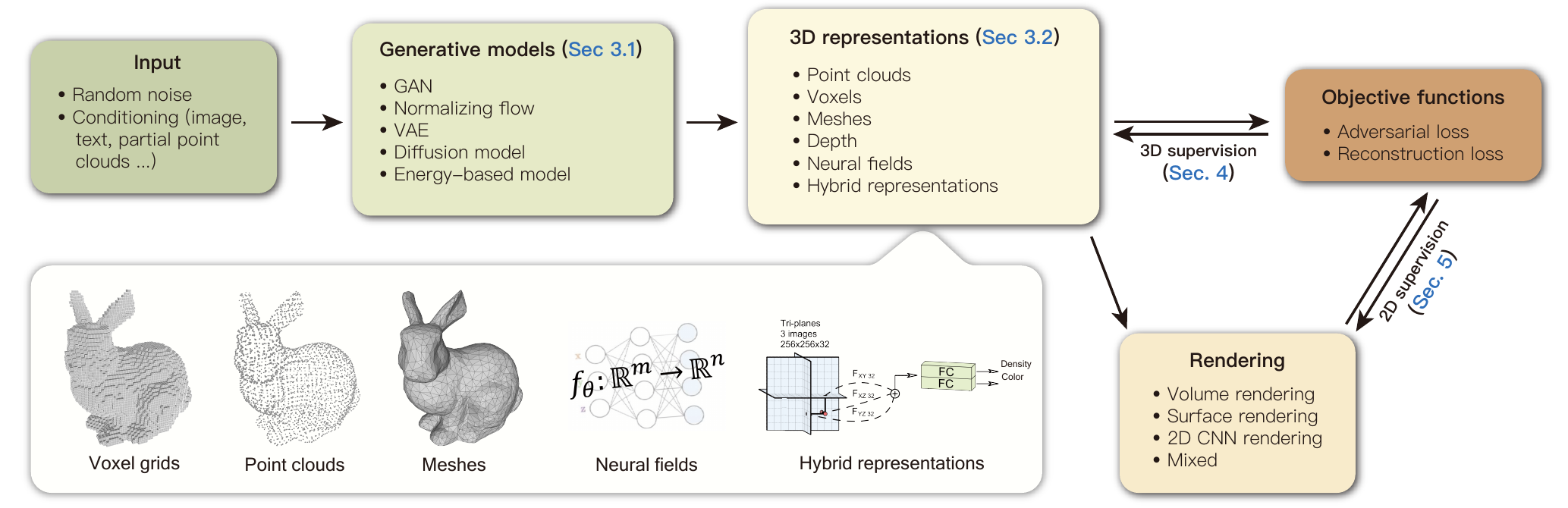}
    \caption{\textbf{3D generative model pipeline.}
    To synthesize 3D data from the random noise or conditioning signal, previous methods propose various types of generative models, such as GAN, normalizing flow, VAE, diffusion model, and energy-based model.
    Popular representations of 3D data include point clouds, voxels, meshes, depth, neural fields, and hybrid representations.
    The generative models are optimized under either the 2D supervision through differentiable rendering or the 3D supervision.
    }
    \label{fig:general_pipeline}
\end{figure*}

%% file: sections/02_scope.tex
\section{Scope of this survey}\label{sec:scope}

This survey primarily focuses on approaches that train networks to model the data distribution of target 3D samples and support sampling for synthesizing 3D representations and 2D images. Additionally, we include methods that predict conditional probability distributions based on specific inputs, such as images, point clouds, or text. It is important to note that these conditional generative methods aim to synthesize 3D representations that respect the inputs while retaining diversity in their generations.
In contrast, classical 3D reconstruction methods establish a one-to-one mapping from inputs to the target 3D representations. Readers seeking a review on such approaches can refer to \cite{neuralrenderingsurvey,3dreconsurvey}.
Furthermore, this survey does not include test-time optimization-based 3D generative methods, such as \cite{jain2022zero, poole2022dreamfusion, lin2023magic3d}. While our survey includes methods that generate 3D representations for rendering, it does not exhaustively cover the field of neural rendering methods, which are thoroughly discussed in \cite{neuralrenderingsurvey,renderingsurvey}.
It is worth mentioning that this survey is complementary to existing surveys on 2D generative models \cite{deepgenerativemodelssurvey,generativemodellingsurvey} and generative models learning from 3D data \cite{Chaudhuri2020CGF}, 
as none of the previous methods provides a comprehensive survey of 3D generative models learning from both 2D and 3D supervision and provides relevant insights.

%% file: sections/03_fundamentals.tex
\section{Fundamentals}\label{sec:fundamentals}

\subsection{Deep Generative Models}

Generative models strive to learn the underlying data distribution in an unsupervised manner, aiming to produce data that is as realistic as possible based on the given information. They enable capturing intricate details and showcasing creativity.
Broadly speaking, generative models can be categorized into two main groups. The first category consists of likelihood-based models, which include variational autoencoders (VAEs)\cite{vae}, normalizing flows (NFs)\cite{normalizingflows}, diffusion probabilistic models (DPMs)\cite{DDPM}, and energy-based models (EBMs)\cite{energybased}. These models learn by maximizing the likelihood of the provided data, allowing them to capture the underlying distribution.
The second category encompasses likelihood-free models, prominently Generative Adversarial Networks (GANs)~\cite{gan}. GANs employ a two-player min-max game framework to find a Nash equilibrium, leading to the generation of synthetic data.
In the subsequent sections, we will provide a concise overview of various generative models, including their characteristics and mechanisms.

\noindent\textbf{Generative Adversarial Networks.} 

Generative Adversarial Networks (GANs), commonly referred to as GANs, have gained immense popularity for their exceptional performance in data synthesis tasks.
Typically, a GAN consists of two separate networks: a generator $G(\cdot)$ and a discriminator $D(\cdot)$. The generator network takes as input a latent code sampled from a prior distribution $\z \sim p_{\z}$ for synthesizing data. On the other hand, the discriminator network aims to distinguish between real data samples $\x \sim p_{\x}$ and the synthesized data $G(\z)$.
During the training process, the generator network $G(\cdot)$ strives to synthesize data that appear as realistic as possible, attempting to deceive the discriminator network $D(\cdot)$ into classifying the generated samples as real. Simultaneously, the discriminator network is trained to accurately label synthesized samples generated by $G(\cdot)$ as fake and training samples $\x$ as real.
These two networks compete with each other and can be formulated into a min-max game.
They are jointly optimized with 
\begin{align}
    \Loss_D & = - \E_{\x\sim p_{\x}} [\log(D(\x))] - \E_{\z\sim p_{\z}}[\log(1-D(G(\z)))], \label{eqa:loss_d} \\
    \Loss_G & = - \E_{\z\sim p_{\z}}[\log(D(G(\z)))]. \label{eqa:loss_g}
\end{align}
With the advent of deep learning, the two networks within GANs have progressively been parameterized using deep neural networks, such as DC-GAN\cite{radford2015unsupervised}.
In recent years, GAN variants like PG-GAN\cite{karras2017progressive} and StyleGAN1-3\cite{stylegan,stylegan2,stylegan3} have emerged, introducing improved architectures capable of synthesizing highly realistic samples.
However, due to the inherent challenges in optimizing the two-player game, GANs often face difficulties in training in a stable manner, which can lead to non-convergence issues. Additionally, GANs are prone to a phenomenon known as mode collapse, where the generator maps multiple distinct latent codes to the same output, resulting in a lack of diversity in the generated samples.

\noindent\textbf{Variational Autoencoders.}

Deep latent variable models (DLVMs) utilize neural networks to parameterize the data distribution, denoted as $\x \sim p_\theta$, by incorporating latent variables $\z$ sampled from a prior distribution $\z \sim p_{\z}$.
However, optimizing the desired parameter $\theta$ of DLVMs presents challenges due to the intractability of the likelihood term $p_\theta(\x) = \int p_{\theta}(\x|\z)p_{\z}(\z)d\z$. This intractability makes differentiation and optimization difficult.
Variational Autoencoders (VAEs) address this issue by transforming the problem of intractable posterior inference into a tractable one. VAEs achieve this by employing an efficient approximation, denoted as $q_{\phi}(\z|\x)$, to the intractable posterior distribution. This approximation enables effective learning and inference within the VAE framework.
Concretely,  $q_{\phi}(\z|\x)$ is parameterized with a feed-forward model and optimized by minimizing the KL divergence between itself and $p_\theta(\z|\x)$:
\begin{align}
    D_{KL}(q_{\phi}(\z|\x)   || p_\theta(\z|\x)) =& \mathrm{log}(p_\theta(\x)) +  D_{KL}(q_{\phi}(\z|\x) ||  p_\theta(\z)) \nonumber \\ 
    &- \E_{\z\sim q_{\phi}(\z|\x)}\mathrm{log}(p_\theta(\x|\z)).
\end{align}
The log likelihood $\mathrm{log}(p_\theta(\x))$ can be re-written into the following form:
\begin{align}
    \mathrm{log}(p_\theta(\x)) = & D_{KL}(q_{\phi}(\z|\x)  || p_\theta(\z|\x)) - D_{KL}(q_{\phi}(\z|\x) || p_\theta(\z)) \nonumber \\
    &+ \E_{\z\sim q_{\phi(\z|\x)}}\mathrm{log}(p_\theta(\x|\z)) \nonumber \\  
    \geq & - D_{KL}(q_{\phi}(\z|\x) || p_\theta(\z)) + \E_{\z\sim q_{\phi(\z|\x)}}\mathrm{log}(p_\theta(\x|\z)),  
\end{align}
where the term $D_{KL}(q_{\phi}(\z|\x) || p_\theta(\z|\x))$ can be eliminated since KL divergence is always non-negative. 
The loss function of a VAE can be expressed as follows:
\begin{align}
    \Loss_{VAE} = - D_{KL}(q_{\phi}(\z|\x) || p_\theta(\z)) + \E_{\z\sim q_{\phi(\z|\x)}}\mathrm{log}(p_\theta(\x|\z)),
\end{align}
which is commonly known as the evidence lower bound (ELBO)\cite{variationalinference}. 

In order to optimize this objective function, the reparameterization trick was introduced in \cite{vae}. This trick enables the generation of samples $\z\sim q_{\phi}(\z|\x)$ to be differentiable, facilitating the training process. The feed-forward nature of $q_{\phi}(\z|\x)$ allows for efficient inference of new samples.
Furthermore, the training process of Variational Autoencoders (VAEs) is generally stable due to the reconstruction-based loss function. This loss function encourages the VAE to accurately reconstruct the given data, leading to improved performance and faithful reconstructions.
However, VAEs are susceptible to a phenomenon known as posterior collapse, wherein the learned latent space becomes uninformative for reconstructing the input data. This can result in a degradation of the generative capacity of the model.
Additionally, due to the injected noise and the inherent imperfections in the reconstruction process, VAEs tend to generate more blurry samples than those produced by GANs, which are known for their ability to produce sharper and more realistic samples.

\noindent\textbf{Normalizing Flows.}

Both GANs and VAEs utilize parametrized models to implicitly learn the density of data, which prevents them from calculating the exact likelihood function for optimizing model training.
To address this limitation, normalizing flows alleviate the problem by introducing a set of invertible transformation functions. These functions enable transforming a simple distribution, such as a standard normal distribution, into the desired probability distribution of the final output.
Specifically, it starts from a normal distribution, and a set of the invertible function $f_{1:N}(\cdot)$ sequentially transforms the normal distribution to the probability distribution of the final output:
\begin{align}
    \z_{i} = f_{i-1}(\z_{i-1}). 
\end{align}
Owing to the invertible properties of the $f_i$, the probability density function of the new variable $\z_{i}$ can be easily estimated from the last step $\z_{i-1}$:
\begin{align}
    p(\z_{i}) &= p(\z_{i-1}) \lvert \frac{d f_i}{d \z_{i-1}}\rvert^{-1}, \\
    \mathrm{log} p(\z_{i}) &= \mathrm{log} p(\z_{i-1}) - \mathrm{log}\lvert \frac{d f_i}{d \z_{i-1}}\rvert .
\end{align}
By applying the chain rule, we can derive the density of the final output $\z_{N}$ after $N$ transformations as follows:
\begin{align}
    \mathrm{log} p(\z_{N}) &= \mathrm{log} p(\z_{0}) - \sum_{1}^{N}\mathrm{log}\lvert \frac{d f_i}{d \z_{i-1}}\rvert ,
\end{align}
where the full chain consisting of $\z_i$ is commonly known as normalizing flow.
Thanks to their invertible property, normalizing flows offer versatility in tasks such as novel sample generation, latent variable projection, and density value estimation. These flows enable straightforward utilization in various scenarios.

However, it struggles with balancing the parameterized model's capacity and efficiency.

\noindent\textbf{Diffusion Models.}
Diffusion models~\cite{DDPM} are parameterized by a Markov chain, which gradually adds noise to the input data $\x_0$ by a noise schedule $\beta_{1:T}$ with $T$ denoting the time steps.
Theoretically, when $T \rightarrow \infty$, $\x_T$ is a normal Gaussian distribution.  
\begin{align}
    q(\x_t|\x_{t-1}) &= \mathcal{N}(\x_t; \sqrt{1-\beta_t}\x_{t-1}, \beta_t \mathbf{I}), \\
    q(\x_{1:T}|\x_0) &= \prod_{t=1}^{T} q(\x_t|\x_{t-1}).
\end{align}
The reverse of the diffusion process is learned to reconstruct the input by modeling the transition $q(\x_{t-1}|\x_t)$ from the noise to the data.
However, the posterior inference $q(\x_{t-1}|\x_t)$ is intractable. This is resolved using a parametric model $p_\theta$ to model the conditional transition probability, achieved by optimizing the Evidence Lower Bound (ELBO) in a manner similar to Variational Autoencoders (VAEs).
Because of the long Markov chain, diffusion models can synthesize high-quality data and allow for stable training.

However, it is important to note that the inference of new samples in diffusion models can be computationally expensive. The sampling process in diffusion models tends to be slower than GANs and VAEs.

\noindent\textbf{Energy-based model.}
Energy-based models leverage the energy function to model the probability distribution of the data explicitly.
It is built upon a fundamental idea that any probability function can be transformed from an energy function by normalizing its volume:
\begin{align}
    p(\x) = \frac{\mathrm{exp}(-E_{\theta}(\x))}{\int_{\x}\mathrm{exp}(-E_{\theta}(\x))},
\end{align}
where $-E_{\theta}(\x)$ is the energy function. 
Clearly, data points with a high likelihood correspond to a low energy value, whereas data points with a low likelihood exhibit a high energy value.
However, it is difficult to optimize the likelihood because calculating $\int_{\x}\mathrm{exp}(-E(\x))$ for high-dimensional data is intractable.
Contrastive Divergence is proposed as a means to mitigate optimization challenges by comparing the likelihood gradient on the true data distribution $p(\x)$
with the gradient on randomly sampled data from the energy distribution $q_\theta(\x)$.

\begin{align}
 \nabla_{\theta} \E_{\x\sim q_\theta}(-\mathrm{log}(p(\x)))  
 =  \E_{\x \sim p}((E_{\theta}(\x))) - \E_{\x\sim q_\theta}((E_{\theta}(\x)))
\end{align}
The energy distribution $q_\theta(\x)$
 
is approximated by Markov Chain Monte Carlo (MCMC) process.

\input{tables/representation_comparison}

\subsection{3D Representations}

The computer vision and computer graphics communities have developed diverse 3D scene representations, such as voxel grids, point clouds, meshes, and neural fields. Each of these representations has its own advantages and limitations when it comes to the task of 3D generation.

In the subsequent sections, we will present the formulations of widely used 3D representations along with their notable works. This background information will lay the foundation for a comprehensive analysis of these representations in the context of 3D generation tasks.
An overview of these 3D representations can be found in Figure \ref{fig:general_pipeline}, providing a visual depiction of their characteristics. Furthermore, a comparison of them for 3D generation with regard to time efficiency, memory efficiency, representation capability, and neural network compatibility is shown in \cref{tab:representation_comparison}.

\noindent\textbf{Voxel grids.} 
Voxels are Euclidean-structured data that is regularly placed in 3D space, akin to pixels in 2D space.
They serve as a representation for 3D shapes and can store various types of information, such as geometry occupancies \cite{3dshapenet,voxnet}, volume densities \cite{xu2021volumegan, schwarz2022voxgraf}, or signed distance values \cite{mittal2022autosdf}.

Thanks to the regularity of voxel grids, they work well with standard convolutional neural networks and are widely used in deep geometry learning.
As a pioneer, 3D ShapeNets~\cite{3dshapenet} introduces voxel grids into 3D scene understanding tasks.
It converts a depth map into a 3D voxel grid, which is further processed by a convolutional deep belief network.
3D-R2N2 \cite{choy20163d}, differently, uses 2D CNNs to encode the input image into the latent vector and utilizes a 3D convolutional neural network to predict the target voxel grid. 
Although voxel grids are well suited to 3D CNNs, processing voxels with neural networks is typically memory inefficient.
Hence, \cite{HSP,OGN,octnet,o-cnn} introduce the octree data structure for shape modeling. 
Voxel grids also have many applications in rendering tasks.
Early methods \cite{deepvoxels, nguyen2019hologan} store high-dimensional feature vectors at voxels to encode the scene geometry and appearance, which will be interpreted into color images using projection and 2D CNNs.
Neural Volumes \cite{neuralvolumes} uses CNNs to predict RGB-Alpha volumes and synthesizes images with volume rendering techniques \cite{volumerendering}.
Multi-plane image (MPI) \cite{MPI} can be regarded as a variation of voxel grids where the 3D space is partitioned into several depth planes, each associated with an RGB-alpha image. By reducing the number of voxels along the depth dimension, MPI-based methods offer computational advantages by reducing the computational cost to some extent.

\label{sec:fund_point_clouds}
\noindent\textbf{Point clouds.}
A point cloud is an unstructured set of points in 3D space, representing a discretized sampling of a 3D shape's surface. Point clouds are commonly generated as direct outputs from depth sensors, making them widely used in various 3D scene understanding tasks. Depth and normal maps can be considered as special cases of point cloud representation.
Although they are convenient to obtain, the irregularity of point clouds makes them difficult to be processed with existing neural networks that are designed for regular grid data (e.g., images).
Moreover, the underlying 3D shape could be represented by different point clouds due to the sampling variations.
Many methods \cite{pointsetgeneration, pointnet, pointnet++, pointcnn, DGCNN, PV-RCNN} have attempted to effectively and efficiently analyze 3D point clouds.
PointNet \cite{pointnet} leverages MLP networks to extract feature vectors from point sets and summarizes features of all points via max-pooling for point order invariance.
PointNet++ \cite{pointnet++} hierarchically groups the point clouds into several sets and separately processes local point sets with PointNet, which captures the local context of point clouds at multiple levels.
Some methods~\cite{DGCNN, PV-RCNN} reformulate point clouds as other types of data structures (\textit{e.g.}, graphs and sparse voxels) and attempt to exploit neural networks in other fields.
To synthesize images with point clouds, a naive way is storing colors on points and rendering point clouds using point splatting.
Since the rendered images tend to contain holes, \cite{neuralrerendering} uses 2D CNNs to refine images.
Some methods \cite{DSS,pulsar} have developed differentiable renderers for point clouds, which can optimize not only for point positions but also for colors and opacity of points.
To increase the modeling capacity of point clouds, \cite{NHR,pointbasedgraphics,synsin,adop} attempt to anchor high-dimensional feature vectors to points and project them to feature maps for the latter rendering.

\noindent\textbf{Meshes.}
Polygonal meshes are non-Euclidean data that represent shape surfaces with a collection of vertices, edges, and faces.
In contrast to voxels, meshes focus solely on modeling the surfaces of 3D scenes, making them more compact representations. Compared to point clouds, meshes provide explicit connectivity information between surface points, enabling the modeling of relationships among points.
Because of these advantages, polygonal meshes are widely used in traditional computer graphics applications, such as geometry processing, animation, and rendering.
However, applying deep neural networks to meshes is more challenging than to point clouds because mesh edges need to be taken into consideration in addition to vertices.
\cite{surfnet,texturenet} parameterize 3D shape surfaces as 2D geometry images and process geometry images with 2D CNNs.%
With the advancement of graph neural networks, \cite{feastnet, meshcnn, meshautoencoder, splinecnn} propose to regard meshes as graphs.
Generating meshes using networks is equally challenging, mirroring the complexities encountered in mesh analysis. The task necessitates predicting not only the vertex positions but also the underlying topology.
\cite{pixel2mesh,pixel2mesh++,shapeviewpoint,categorymesh} pre-define a mesh with fixed connectivity and predict vertex displacements to deform the mesh to produce the target shape.
In the rendering pipeline of traditional computer graphics, both software and hardware have been heavily optimized for rendering meshes.
Some differentiable mesh renderers \cite{opendr,neural3dmeshrenderer,softrasterizer} leverage the advances of classical rendering techniques and design the back-propagation process to update some properties (\textit{e.g.}, colors) defined on meshes.
To improve the rendering quality, a strategy is storing appearance properties on shape surfaces, which are parameterized as texture maps.
Learning-based methods \cite{deferredneuralrendering,ANR} define learnable feature vectors in texture maps, which are decoded into color images with a 2D renderer.

\input{tables/method_summary}

\noindent\textbf{Neural fields.}
A neural field is a continuous neural implicit representation that encompasses the complete or partial depiction of scenes or objects using a neural network.
For each position in 3D space, the neural network maps its related features (\textit{e.g.}, coordinate) to the attributes (\textit{e.g.}, an RGB value).
Neural fields \cite{onet, deepsdf, unisurf, volsdf, neus, differentiablevolumetricrendering, nerf} are able to represent 3D scenes or objects in arbitrary resolution and unknown or complex topology due to their continuity in representation.
In addition, in comparison to the aforementioned representations, the storage requirements are limited to the parameters of the neural network, resulting in reduced memory consumption when compared to alternative representations.
To render an image from a neural field, there are two streams of techniques - surface rendering and volume rendering.
Surface rendering \cite{spheretracing, takikawa2021neural} utilizes an implicit differentiable renderer to trace viewing rays and determine their intersection points with the surface. Subsequently, the network is queried to obtain the RGB values associated with these intersection points, which are then used to generate a 2D image.
While surface rendering-based methods demonstrate strong performance in representing 3D objects and rendering 2D images, they often necessitate per-pixel object masks and meticulous initialization to facilitate optimization towards a valid surface. This requirement arises from the fact that surface rendering solely offers gradients at the points where rays intersect with the surface, posing challenges in network optimization.
Volume rendering~\cite{volumerendering}, in contrast, is based on ray casting and samples multiple points along each ray. It has shown great power in modeling complex scenes.
NeRF~\cite{nerf} and its following works adopt such differentiable volume rendering to render 2D images from 3D representations, allowing gradients to propagate through the renderer.
However, sampling a set of points along all rays may lead to low rendering speed. Recent works focus on acceleration via various techniques, such as pruning~\cite{neuralsparsevoxel}, improved integration~\cite{autointegration}, and carefully-designed data structures~\cite{plenoctrees,baking,fastnerf,kilonerf}.

\noindent\textbf{Hybrid representation.}\label{sec:hybrid representation}
Given the respective advantages and disadvantages of each representation, hybrid representations have been proposed as a means to complement and combine their strengths.
Many of these hybrid representations primarily concentrate on the fusion of explicit and implicit representations.
Explicit representations provide explicit control over the geometry. On the other hand, they are restricted by the resolution and topology.
Implicit representations allow for the modeling of complex geometry and topology with relatively low memory consumption. However, they are usually parameterized with MLP layers and output the attribute for each coordinate, suffering from small receptive fields. Consequently, explicit supervision on surfaces is challenging, and optimization becomes difficult.
Researchers leverage the advantages of each representation type to compensate for the drawbacks of the other.
Some works~\cite{neuralsparsevoxel,baking,plenoxels} integrate voxel grids into neural fields to accelerate the training and rendering processes. The features of points for differentiable rendering are interpolated from features of voxel grids.
These representations sacrifice memory consumption for rendering speed.
MINE~\cite{mine} merges neural fields with multi-plane images, which is a smaller representation than voxel grids but suffers from a limited viewing range. 
EG3D~\cite{eg3d} uses tri-planes to boost the model capacity of neural fields. 
Such representations consume less memory than voxel-based neural fields, and they allow fast rendering at the same time.
\cite{point-nerf} and \cite{points2nerf} build a neural field on a point cloud.
\cite{point-nerf} interpolates point features from K neighboring points, while \cite{points2nerf} uses a hyper-network that takes in a point cloud and then generates the weights of NeRF network.
NeuMesh~\cite{neumesh} presents mesh-based neural fields by encoding geometry and texture codes on mesh vertices, enabling the manipulation of neural fields through meshes.
\cite{tetrahedralmeshes}, in contrast, combines two explicit representations, mesh and voxel grids. The proposed deformable tetrahedral mesh representation optimizes both vertex placement and occupancy. 
This representation achieves both memory and computation efficiency.

%% file: tables/representation_comparison.tex
\begin{table*}
    \centering
    \SetTblrInner{rowsep=3pt}       %
    \SetTblrInner{colsep=5pt}       %
    \caption{%
        \textbf{Comparison of different representations} with regard to time efficiency, memory efficiency, representation capability, and NN (neural network) compatibility. A larger number of stars indicates better performance.
    }
    \label{tab:representation_comparison}
    \begin{tblr}{
        cells={halign=c,valign=m},  %
        hline{1,2,6}={1-7}{},   %
        hline{1,6}={1.0pt},        %
        hline{2}={0.7pt},           %
        vline{2,3,4,5,6,7}={1-5}{},      %
    }
        \textbf{3D Representation}  & \raisebox{-.15\height}{\includegraphics[width=0.3cm]{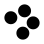}}\textbf{Point Clouds}  & \raisebox{-.15\height}{\includegraphics[width=0.3cm]{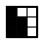}}\textbf{Voxel Grids}   & \raisebox{-.15\height}{\includegraphics[width=0.3cm]{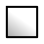}}\textbf{Depth/Normal Maps} & \raisebox{-.15\height}{\includegraphics[width=0.3cm]{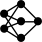}}\textbf{Neural Fields} & \raisebox{-.15\height}{\includegraphics[width=0.3cm]{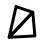}}\textbf{Mesh}  & \raisebox{-.15\height}{\includegraphics[width=0.3cm]{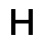}}\textbf{Hybrid} \\
        Time Efficiency   & $\star\star\star$           & $\star$             & $\star\star\star\star\star$             &   $\star$          & $\star\star\star\ \star$  & $\star\star\star$   \\
  Memory Efficiency & $\star\ \star$            & $\star\ \star$            & $\star\star\star\ \star$              &   $\star\star\star\star\star$       & $\star\star\star$   & $\star\star\star$\\
  Representation Capability&  $\star\star\star$   & $\star\ \star$            &   $\star$               &     $\star\star\star\star\star$     & $\star\star\star\ \star$  &  $\star\star\star\star\star$  \\
  NN Compatibility & $\star\ \star$             & $\star\star\star\star\star$         & $\star\star\star\star\star$             &   $\star\star\star\star\star$       & $\star$     & $\star\star\star\star\star$ \\
    \end{tblr}
\end{table*}

%% file: tables/method_summary.tex
\begin{table*}[t]
\centering
\caption{\textbf{Properties of representative 3D generative models.} Units for ``FLOPS" and ``\#Params." are in G and M, respectively. We calculate the two metrics for methods implemented using a PyTorch~\cite{NEURIPS2019pytorch} implementation. }
\label{tab:properties}

\begin{subtable}{0.49\textwidth}
\caption{Methods using 3D supervision.}
\label{tab:shape_generation}
\centering\fontsize{6pt}{6pt}\selectfont
\SetTblrInner{rowsep=3.0pt}       %
\SetTblrInner{colsep=3.5pt}       %
\begin{tblr}{
    cells={halign=c,valign=m},  %
    hline{1,2,15}={1-5}{},   %
    hline{1,15}={1.0pt},        %
    hline{2}={0.7pt},           %
    row{odd} = {gray!0}, 
    row{even} = {gray!5},
}
\textbf{Method} & \textbf{3D Representation} & \textbf{FLOPS} & \textbf{\#Params.} & \textbf{Shape Resolution}\\
PQ-Net~\cite{wu2020pq} 
& Voxel grid
& 0.02
& 13.62
& $64\times64\times64$
\\
AutoSDF~\cite{mittal2022autosdf} 
& Voxel grid
& 997.41
& 2941.10
& $64\times64\times64$
\\
PointFlow~\cite{yang2019pointflow} 
& Point cloud 
& 105.89
& 1.06
& 2048
\\
Hui~\etal~\cite{hui2020progressive} 
& Point cloud 
& 25.79
& 12.71
& 2048
\\
ShapeGF~\cite{cai2020learning} 
& Point cloud 
& 413.85
& 4.17
& 2048
\\
SoftFlow~\cite{kim2020softflow} 
& Point cloud 
& 5.26
& 6.50
& 2048
\\
SP-GAN~\cite{li2021sp} 
& Point cloud 
& 1.64
& 0.59
& 2048
\\
Generative PointNet~\cite{xie2021generative} 
& Point cloud 
& 91.86
& 1.39
& 2048
\\
Luo~\etal~\cite{luo2021diffusion} 
& Point cloud 
& 67.33
& 3.30
& 2048
\\
IM-Net~\cite{chen2019learning} 
& Neural field
& 799.70
& 3.05
& $256 \times 256 \times 256$
\\
Deng~\etal~\cite{deng2021deformed} 
& Neural field 
& 209.11
& 0.10
& $256 \times 256 \times 256$ 
\\
gDNA~\cite{chen2022gdna} 
& Neural field 
& 425.90
& 0.87
& $256 \times 256 \times 256$
\\
ShapeAssembly~\cite{jones2020shapeassembly} 
& Program 
& 0.02
& 1.41
& -  
\end{tblr}
\end{subtable}
\hfill
\begin{subtable}{0.49\textwidth}
\caption{Methods using 2D supervision.}
\label{tab:3dimage_generation}
\centering\fontsize{6pt}{6pt}\selectfont
\SetTblrInner{rowsep=3.0pt}       %
\SetTblrInner{colsep=4.8pt}       %
\begin{tblr}{
    cells={halign=c,valign=m},  %
    hline{1,2,15}={1-5}{},   %
    hline{1,15}={1.0pt},        %
    hline{2}={0.7pt},           %
    row{odd} = {gray!0}, 
    row{even} = {gray!5},
}
\textbf{Method}  & \textbf{3D Representation} & \textbf{FLOPS} & \textbf{\#Params.} & \textbf{Image Resolution} \\
Shi~\etal~\cite{shi20223d}
& Depth map
& 63.24
& 55.81
& $256 \times 256$
\\
Schwarz~\etal~\cite{graf}
& Neural field
& 177.03
& 0.68
& $64 \times 64$
\\
Niemeyer~\etal~\cite{giraffe}
& Neural field
& 5.21
& 0.75
& $256 \times 256$
\\
Chan~\etal~\cite{pigan}
& Neural field
& 842.42
& 1.91
& $256 \times 256$
\\
Pan~\etal~\cite{shadegan}
& Neural field
& 1243.66
& 1.91
& $256 \times 256$
\\
Niemeyer~\etal~\cite{campari}
& Neural field
& 183.86
& 0.36
& 128 $\times$ 128
\\
Xu~\etal~\cite{generativeoccupancy}
& Neural field
& 621.53
& 1.91
& $256 \times 256$
\\
Sun~\etal~\cite{fenerf}
& Neural field
& 2094.73
& 2.70
& 256 $\times$ 256
\\
StyleNeRF~\cite{stylenerf}
& Neural field
& 13.49
& 7.63
& $1024 \times 1024$
\\
GRAM~\cite{gram}
& Neural field
& 977.91
& 1.95
& $256 \times 256$
\\
GIRAFFE HD~\cite{xue2022giraffehd}
& Neural field
& 35366.12
& 12.78
& 1024 $\times$ 1024
\\
VolumeGAN~\cite{xu2021volumegan}
& Hybrid
& 32.52
& 8.67
& $256 \times 256$
\\
EG3D~\cite{eg3d}
& Hybrid
& 25.78
& 30.60
& 512$\times$512
\end{tblr}
\end{subtable}

\end{table*}

%% file: sections/04_shape_generation.tex
\input{figures/3d_generator.tex}

\section{Learning from 3D data}\label{sec:shape_generation}

With the availability of 3D data, a majority of recent 3D generative methods focus on training deep neural networks to effectively capture the distributions of 3D shapes. Unlike 2D images, 3D shapes can be represented in various ways, such as voxel grids, point clouds, meshes, and neural fields. Each of these representations possesses distinct advantages and disadvantages when applied to the task of 3D generation from 3D data.
Several factors come into play when evaluating the compatibility of a 3D representation with deep generative models. These factors include the ease of network processing for a given representation, the ability to efficiently generate high-quality and intricate 3D shapes, and the costs associated with obtaining supervision signals for the generative models. Assessing these aspects is crucial to determine the suitability of a particular 3D representation for successful integration with deep generative models.
\cref{fig:3d_generator} summarizes the representative pipelines on 3D-supervision-based methods, and \cref{tab:representation_classification} outlines the representative methods. \cref{tab:shape_generation} summarizes the properties of the representative methods.

\subsection{Voxel Grids}

Voxel grids are usually seen as images in 3D space. To represent 3D shapes, voxels could store the geometry occupancy, signed distance values, or density values, which are implicit surface representations that define the shape surface as a level-set function.
Thanks to the regularity of its data structure, the voxel grid is one of the earliest representations being used in deep learning techniques for 3D vision tasks, such as 3D classification \cite{voxnet,3dshapenet,brock2016generative}, 3D object detection \cite{voxelnet,voxeltransformer}, and 3D segmentation \cite{voxelembed,vvnet}.

Wu \etal \cite{wu2016learning} adopt the architecture of generative adversarial networks to process 3D voxel grids.
The generator maps a high-dimensional latent vector to a 3D cube, which describes the synthesized object in voxel space.
In contrast to 2D GANs, the generator and discriminator in 3D GANs are constructed using a sequence of 3D convolution blocks to enable the processing of 3D voxel data.
In practice, they construct an encoder to map the 2D image into the latent space of its corresponding generator, closely resembling the VAE-GAN~\cite{larsen2016autoencoding}.
In addition to the conventional adversarial loss, they incorporate two components for encoder training: a reconstruction loss and a KL divergence loss. These additional elements serve to constrain the output distribution of the encoder.
Owing to its VAE-GAN-like design, the proposed model can also be leveraged to recover 3D shapes from the 2D observation.
They also demonstrate that the discriminator learned without any supervision can be successfully transferred to several 3D downstream tasks with good performance. 
Considering the training of GAN models is unstable, \cite{brock2016generative} attempts to train a variational auto-encoder to model the shape distribution.
It first uses an encoder network consisting of four 3D convolutional layers and a fully connected layer to map the input voxel into a latent vector and then the 3D decoder with an identical but inverted architecture to transform a latent vector into a 3D voxel.   
The downsampler in the encoder and upsampler in the decoder are implemented by stride convolutions. 
The objective function for training consists of two parts: one is KL divergence on the latent codes, and the other is a Binary Cross-Entropy (BCE) for voxel reconstruction.
They incorporate modifications to the BCE loss to avoid gradient vanishing.
Despite its ability to handle dense objects, this method exhibits a limitation in generating smooth rounded edges, akin to the behavior observed in 2D VAEs, which often leads to the generation of blurry voxels.
To solve these problems, \cite{mittal2022autosdf} introduces VQ-VAE to model the data distribution. 
Different from VAE, only using a single vector serving as a latent for input,
they use VQ-VAE to project the high-dimensional 3D shape into a lower-dimensional discrete latent space which is optimized during training, not fixed like VAE.
Once the latent space is well trained, they use a transformer to autoregressively model the non-sequential data. 
Specifically, they maximize the likelihood of the latent representations using randomized orders for an autoregressive generation. 
A well-trained autoregressive model holds the potential for diverse downstream applications, such as shape completion and text-guided shape generation. The incorporation of conditional information into the autoregressive model enables straightforward fusion with the model, facilitating these applications efficiently.

In addition to modeling the overall shape of the object, some works endeavor aim to achieve more detailed and fine-grained shape generation.
SAGNet \cite{wu2019sagnet} proposes to use an autoencoder to jointly learn the geometry of the part and the pairwise relationship between different parts.
It uses a two-way encoder to independently extracts the features of both geometry and structure.
GRU and attention modules are also incorporated into the encoder to exchange the geometric and structural information encoded by two independent encoders and summarize the input into a latent code.
The architectural design allows for the disentangled control of the object structure.
Li \etal \cite{li2020learning} also propose to model 3D shape variations at the part level. In addition, they explore the automated assembly of various parts to form a complete 3D shape.
Specifically, they first learn a part-wise generative network that consists of K part generators, where K is the number of parts.
The part generator is built upon 3D VAE-GAN, which is very similar to \cite{wu2016learning}.
The difference is that Li \etal \cite{li2020learning} additionally introduce a reflective symmetry loss to encourage the symmetry property of the generated object.
To assemble the synthesized parts with different scales and positions, they propose a part assembler to regress the transformation matrices for each part.
Since the assembling solution is not unique for the given parts, they define a fixed anchor part while the remains are required to learn the transformation matrices to match the anchor part.
PQ-Net \cite{wu2020pq} designs a sequence-to-sequence (Seq2Seq) network for the part assembly.
Thanks to the Seq2Seq modeling, its network demonstrates impressive performance for several tasks, including shape generation and single-view 3D reconstruction.

Although the above approaches perform well on low-resolution voxel grids, they struggle to handle high-resolution voxel grids that contain fine-grained details due to the cubic growth in computational complexity.
To alleviate these issues, Ibing \etal \cite{ibing2021octree} attempt to use the octree as a hierarchical compact representation for the voxel grid, where they convert the octree into a sequence by the traversal order.
Besides, an adaptive compression scheme is used to decrease the sequence length to improve the generation efficiency.

On the one hand, similar to 2D images, voxel grids are in an Euclidean structure, which works well with 3D CNNs. 
On the other hand, the voxel grid typically consumes much computation cost because the number of voxel elements grows cubically with resolution.
Although the octree for voxel grids can reduce a lot of computational cost, it cannot be processed by neural networks very efficiently due to its non-grid structure.

\subsection{Point Clouds}

Since point clouds are the direct outputs of depth scanners, they are widely used in scene understanding tasks.
Leveraging generative models to model data priors of specific point cloud datasets can benefit various downstream computer vision tasks \cite{li2018point, stylepointcloudcompletion}.
In contrast to voxel grids, point clouds are an explicit surface representation and directly characterize shape properties, which has two advantages. First, deep neural networks usually process point clouds with less GPU memory.
Second, they are suitable for some generative models, such as normalizing flows and diffusion models.
Despite the two advantages, the irregularity of point clouds makes networks difficult to analyze and generate them.

As the pioneer work, Achlioptas \etal \cite{achlioptas2018learning} exploit generative adversarial networks to learn the distributions of 3D point clouds. 
It proposes a raw point cloud GAN (r-GAN) and a latent-space GAN (l-GAN). The generator of the r-GAN is an MLP network with 5 fully connected layers, which maps the randomly sampled noise vector to the point cloud with 2048 points. The corresponding discriminator uses PointNet \cite{pointnet} as the network backbone.
Achlioptas \etal \cite{achlioptas2018learning} found that r-GAN has difficulty in generating high-quality point clouds. A plausible reason is that GANs are hard to converge to a good point.
To overcome this problem, they present a novel training framework that trains a generative adversarial network to model the latent space of a pre-trained auto-encoder, which is called l-GAN.
The l-GAN delivers much better performance than r-GAN.
The GAN generator of point clouds is typically implemented as fully-connected networks, which cannot effectively leverage local contexts for producing point clouds.
To solve this problem, some methods \cite{valsesia2018learning, shu20193d, arshad2020progressive, hui2020progressive} propose to construct the GAN generator based on the graph convolution.
For example, given a sampled latent vector, Valsesia \etal \cite{valsesia2018learning} first use an MLP network to predict a set of point features, which is taken as a graph and processed by a graph convolutional network.
When upsampling the point cloud, it applies the graph convolution to point features to obtain new feature vectors, which are concatenated to the original point features to produce the upsampled point set.
Another challenge of point cloud generation is that synthesizing high-resolution point clouds easily consumes a lot of memory.
Ramasinghe \etal \cite{ramasinghe2020spectral} reduce the computational complexity by adopting a GAN model in the spectral domain.
It represents point clouds as spherical harmonic moment vectors and regresses these vectors from sampled latent vectors with MLP networks.

While GAN-based methods have demonstrated remarkable generation performance, the inherent instability in their training process has prompted researchers to investigate alternative types of generative models.
Zamorski \etal \cite{zamorski2020adversarial} extend the variational auto-encoder model (VAE) and adversarial auto-encoder model (AAE) to the 3D domain.
The encoder is implemented as a PointNet-like network, and the decoder is an MLP network.
Since point clouds are irregular, the reconstruction loss for the auto-encoder is implemented as set-to-set matching loss, such as Earth Mover's distance and Chamfer distance.
Similar to the 2D VAE model, the 3D VAE model also employs the KL divergence to supervise the latent space.
Given the potential intractability of the KL divergence, the AAE model adopts an alternative approach by learning the latent space through adversarial training.
Zamorski \etal \cite{zamorski2020adversarial} empirically find that the 3D AAE model outperforms the 3D VAE model in terms of performance.

Due to the non-Euclidean data structure of point clouds, the GAN-based and AE-based generative models mostly use MLP networks or graph convolutional networks to map latent vectors to point clouds, which can typically produce a fixed number of points.
This significantly limits their modeling ability. 
Even for shapes within the same category, their complexity could require different numbers of points.
To overcome this problem, PointFlow \cite{yang2019pointflow} models point clouds as a distribution of distributions and introduces a normalizing flow model to generate point clouds.
Specifically, PointFlow first samples a set of points from a generic prior distribution, such as standard Gaussian.
Then, it samples a latent vector from the shape distribution that encodes the shape information and feeds the vector into a conditional continuous normalizing flow (CNF), which produces a vector field to move the sampled points to generate the shape.
PointFlow assumes that modeling shape prior as the Gaussian distribution limits the performance of VAE models and thus uses an additional CNF to model the shape prior.
Since the Neural ODE solver in continuous normalizing flow is computationally expensive, Klokov \etal \cite{klokov2020discrete} propose to adopt the affine coupling layers to build discrete normalizing flows, resulting in a significant speedup.
Another challenge encountered in flow-based models is their potential failure when the dimensions of the data and target distributions do not match.
The point clouds typically lie on 2D manifolds, while a generic prior distribution is defined over the 3D space, making flow-based models struggle to transform point clouds to match the prior distribution.
To solve this issue, SoftFlow \cite{kim2020softflow} perturbs point clouds with sampled noise and uses a conditional normalizing flow model to map perturbed points to the latent variables.

Recent methods \cite{cai2020learning, luo2021diffusion, zhou20213d,zeng2022lion,wu2022fast} model the point cloud generation as a denoising process and train a model to output vector fields to gradually move points from a generic prior distribution.
ShapeGF \cite{cai2020learning} regards the shape as a distribution and assumes that points on the shape surface have high densities.
For any 3D point, it trains a network to predict the point's gradient that is used to move the point to the high-density area.
\cite{luo2021diffusion, zhou20213d,zeng2022lion,wu2022fast} formulate the generation of point clouds as a reverse diffusion process, which is in contrast to the diffusion process in non-equilibrium thermodynamics.
The reverse diffusion process is implemented as a Markov chain that transforms the distribution of points from the noise distribution to the target distribution.
Each transition step is instantiated as an auto-encoder, which takes the point cloud as input and outputs the displacements of points.

PointGrow \cite{sun2020pointgrow} develops an auto-regressive model that recurrently predicts new points conditioned on previously generated points.
It designs self-attention modules to capture long-range dependencies.
However, the order of points is hard to define.
PointGrow \cite{sun2020pointgrow} sorts points based on the points' z coordinates.
Xie \etal \cite{xie2021generative} propose a deep energy-based model for synthesizing point clouds.
The short-run MCMC is adopted as the point generator.
The energy model is implemented as a PointNet-like network, which predicts scores for generated point clouds.

Recently, some methods \cite{gal2020mrgan, li2021sp} attempt to attain part-level controllability over the generated shapes.
MRGAN \cite{gal2020mrgan} proposes a multi-root GAN that consists of multiple branches, and each branch map a sampled latent vector to a set of points, which are concatenated into the final point cloud.
After training, the parts of generated objects can be edited by revising the corresponding latent vectors.
SP-GAN \cite{li2021sp} adopt the sphere as prior for the part-aware shape generation.
It defines a fixed point cloud on the unit sphere and anchors the sampled latent vector to each point, which is then fed into a generator to produce the shape.
The initial point cloud acts as a guidance to the generative process and provides dense correspondences between generated shapes, naturally enabling part-level controllability.
Changing latent vectors of some points leads to the modification of the associated part.

Point clouds have been adopted in many types of generative models to synthesize shapes.
Although previous methods have achieved impressive performance in shape generation, high-resolution shapes are still difficult to obtain.
The reason is that modeling high-resolution shapes requires a significant number of points, which will consume a large amount of GPU memory.

\subsection{Neural Fields}

Neural fields use neural networks to predict properties for any point in the 3D space.
Most of them adopt MLP networks to parameterize 3D scenes and can model shapes of arbitrary spatial resolutions in theory, which is more memory efficient than voxel grids and point clouds.
Although neural representations have superiority in shape modeling, it is not straightforward to apply common generative models such as GANs on these representations, due to the lack of ground-truth data in neural representation format and the difficulty of processing neural representations with neural networks directly.
To overcome this problem, some methods use auto-decoders~\cite{autodecoder} to model the distribution.
gDNA~\cite{gdna} adopts an auto-decoder for dynamic human generation. 
A 3D CNN-based feature generator first processes the shape latent code into a feature volume, which is further decoded into occupancy values and feature vectors through an MLP network.
Deng \etal \cite{deng2021deformed} aim to preserve shape correspondences when generating shapes. 
An MLP network is used to represent a template signed distance field that is shared among all instances.
The deformation and correction fields are modeled by another two MLPs in the template space.
DualSDF \cite{hao2020dualsdf} learns a shared latent space to enable semantic-aware shape manipulation. 
A sampled latent code will be processed by two networks that handle different levels of granularity.
One is with SDF that can capture fine details, while the other one is with simple shape primitives to represent a coarse shape.
Two reconstruction losses are calculated between the given shapes and the generated shapes of two representations.

To generate implicit fields based on generative adversarial networks, some methods discriminate the generated implicit fields either in the latent space or with the converted explicit representations.
Chen \etal \cite{chen2019learning} apply the discriminator on the latent space. 
An auto-encoder is first used to learn the features from a set of shapes, where the encoder can be a 3D CNN or PointNet~\cite{pointnet}, and the decoder is parameterized with an MLP network.
Then, latent-GANs~\cite{achlioptas2018learning, arjovsky2017towards} are employed to train on the features extracted by the pre-trained encoder.
Ibing \etal \cite{ibing20213d} also leverage latent-GANs but propose a hybrid representation as the intermediate feature for learning to enable spatial control of the generated shapes.
The latent representation combines voxel grids and implicit fields, and therefore each cell covers a part of the shape.
Kleineberg \etal \cite{kleineberg2020adversarial} generate signed distance fields and design two types of discriminators, voxel-based (\textit{e.g., 3D CNN}) and point-based (\textit{e.g.}, PointNet), for the training.
In voxel-based cases, a fixed number of points are fed into the generator to query signed distance values.
While in point-based cases, points of arbitrary sequence can be queried for signed distance values.
SurfGen~\cite{surfgen} develops a differentiable operator that extracts surface from implicit fields through marching cubes~\cite{marchingcubes} and then performs differentiable spherical projection on the surface, which is an explicit shape representation.
A spherical discriminator operates on explicit surfaces to supervise the learning of the shape generator.

With the development of diffusion models, it is interesting to apply powerful generative models to learn the distributions of neural fields.
\cite{dupont2022data} starts the initial attempt by representing data using an implicit neural representation and learning a diffusion model directly on the modulation weights of the implicit function.
However, blurry results are generated compared to the results from GAN-based methods. 
\cite{chou2022diffusionsdf} and \cite{nam20223d} leverage auto-encoders to compress SDF representation to a latent representation and then model the distribution on latent space using diffusion models.
Some works~\cite{wang2022rodin,shue20223d} represent neural fields with triplanes and obtain the triplane representations from multi-view dataset first. Then, diffusion models are adopted to model the distribution of these triplane representations.

\input{tables/representation_classification}

\subsection{Meshes and Other Representations}

Mesh is one of the most used representations in traditional computer graphics.
It is also usually taken as the target object in many 3D modeling and editing softwares.
Despite the mesh's popularity in traditional applications, it is challenging to apply deep generative models to mesh generation due to two factors.
First, meshes are non-Euclidean data and cannot be directly processed by convolutional neural networks.
Second, mesh generation requires synthesizing meshes with plausible connections between mesh vertices, which is difficult to achieve.

To avoid handling the irregular structure of meshes, Ben-Hamu \etal \cite{ben2018multi} propose an image-like representation called multi-chart structure to parameterize the mesh.
They define a set of landmark points on a base mesh, and each triplet of landmark points corresponds to a function that establishes correspondences between a chart on the image domain and the shape surface.
By representing meshes with multi-chart structures, this approach can utilize well-developed image GAN techniques to generate shapes.
Nevertheless, the parametrization trick necessitates a congruent topology between the generated meshes and the base mesh employed in defining the multi-chart structure.
To make the generation process easier, SDM-Net \cite{gao2019sdm} registers training meshes with a unit cube mesh, which enables them to have the same connectivity as the template cube mesh.
SDM-Net utilizes the variational auto-encoder to learn the distributions of meshes, where the encoder and decoder are implemented with convolutional operators defined on meshes.
Based on SDM-Net \cite{gao2019sdm}, TM-Net \cite{gao2021tm} additionally defines a texture space on the template cube mesh and uses a CNN-based VAE to synthesize texture maps, which are combined with generated meshes to produce textured meshes.
PolyGen \cite{nash2020polygen} attempts to synthesize the connectivity of meshes based on an auto-regressive generative model.
It develops a transformer-based network to sequentially generate mesh vertices and faces.
Similar to PointGrow \cite{sun2020pointgrow}, PolyGen sorts the mesh vertices along the vertical axis and leverages a vertex transformer to generate vertices.
Then, mesh faces are predicted using a face transformer that is conditioned on generated mesh vertices.
Liu \etal \cite{liu2023meshdiffusion} parameterize 3D meshes with tetrahedral grids, and each grid is associated with a deformation offset and an SDF value. Such a representation is treated as the input of diffusion models to model the underlying distribution.
Lyu \etal \cite{lyu2023controllable}, in contrast, introduce point clouds as the intermediate representation and leverage point cloud diffusion model for shape generation.

Representing shapes as structured computer programs is an attractive direction, as programs guarantee the production of high-quality geometries and are editable by users.
ShapeAssembly \cite{jones2020shapeassembly} proposes to create programs using a VAE model.
To construct the training dataset, it turns 3D shapes into programs and develops a sequence VAE for learning.
To analyze the input program, the encoder utilizes MLP networks to extract features from each program line and fuse them into a latent vector with a GRU module.
During program generation, the decoder uses the GRU to sequentially predict features for each line, which are then mapped to the program line using MLP networks.

%% file: figures/3d_generator.tex
\begin{figure*}[t]
    \centering
    \includegraphics[width=1\textwidth]{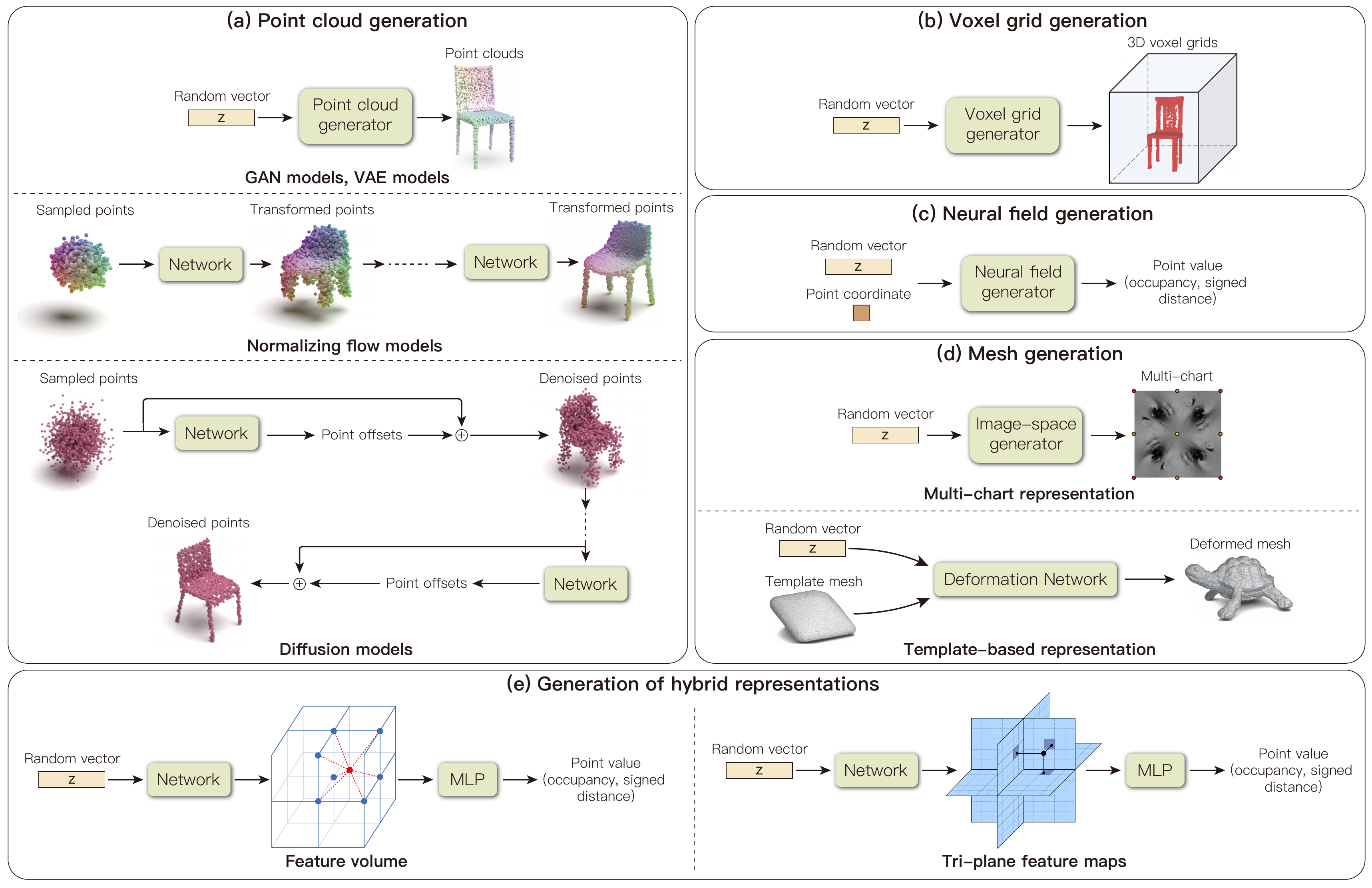}
    \caption{
    \textbf{Representative 3D generative models.}
    We present some classical pipelines for generating (a) point clouds, (b) voxel grids, (c) neural fields, (d) meshes, and (e) hybrid representations.
    Some figures are taken from \cite{li2021sp,yang2019pointflow,luo2021diffusion,wu2016learning,ben2018multi,Hanocka2020p2m,peng2021neural,Gao2022NeurIPS}.
    We only present some of representative methods.
    Please refer to Sec.~\ref{sec:shape_generation} for more variants.
    }
    \label{fig:3d_generator}
\end{figure*}

%% file: tables/representation_classification.tex
\begin{table*}
    \centering\scriptsize
    \SetTblrInner{rowsep=1.0pt}       %
    \SetTblrInner{colsep=1.5pt}       %
    \caption{%
        \textbf{Representative 3D generative models categorized by supervision and 3D representation.}
        Methods in \com{green} model 3D representation in a compositional way.
        \gloseman{Orange} denotes that the method supports controllability through semantic maps, while those in \gloreli{pink} support relighting.
        Methods in \glohuman{blue} enable control through human pose.
        Note that generative models, such as energy-based models and normalizing flows, are not included as this table focuses on the most frequently used types of 3D generative models (GANs, VAEs, and Diffusion Models).
    }
    \label{tab:representation_classification}
    \begin{tblr}{
        cells={halign=c,valign=m},  %
        cell{1}{1}={r=2}{},         %
        cell{1}{2}={r=2}{},         %
        cell{1}{3}={c=3}{},         %
        cell{3}{1}={r=4}{},         %
        cell{7}{1}={r=4}{},
        hline{2}={3-5}{},
        hline{1,3,7,11}={1-5}{},
        hline{4,5,6,8,9,10}={2-5}{},   %
        vline{1,2,3,6}={1-10}{},      %
        vline{4,5}={2-10}{},      %
        column{3}={7.85cm},
        column{4}={2.4cm},
        column{5}={3.5cm},
    }
     \textbf{Supervision} & \textbf{3D Representation} & \textbf{Generative Model}    &               &    \\
                 &                   &  \textbf{GAN}               & \textbf{VAE}           & \textbf{Diffusion} \\
3D   & Point Clouds      & \glo{Achlioptas \etal\cite{achlioptas2018learning}}, \glo{Shu \etal\cite{shu20193d}}, \glo{Valsesia \etal\cite{valsesia2018learning}}, \glo{Spectral-GAN\cite{ramasinghe2020spectral}}, \glo{Hui \etal\cite{hui2020progressive}}, \glo{Arshad \etal\cite{arshad2020progressive}}, \com{SP-GAN\cite{li2021sp}}, \com{MRGAN\cite{gal2020mrgan}} & \glo{Zamorski \etal\cite{zamorski2020adversarial}} & \glo{ShapeGF\cite{cai2020learning}}, \glo{Luo \etal\cite{ luo2021diffusion}}, \glo{PVD\cite{zhou20213d}}, \glo{lion\cite{zeng2022lion}}, \glo{PSF\cite{wu2022fast}} \\
                 & Voxel Grids       & \glo{Wu \etal\cite{wu2016learning}}, \glo{Ibing \etal\cite{ibing2021octree}},\com{SAGNet\cite{wu2019sagnet}}, \com{PQ-Net\cite{wu2020pq}}, \com{PAGENet\cite{li2020learning}} & \glo{AutoSDF\cite{mittal2022autosdf}}, \glo{Brock \etal\cite{brock2016generative}} & \glo{DiffRF\cite{muller2023diffrf}} \\
                 & Neural Fields     & \glo{IM-Net\cite{chen2019learning}}, \glo{Kleineberg \etal\cite{kleineberg2020adversarial}}, \glo{Ibing \etal\cite{ibing20213d}}, \glo{gDNA\cite{gdna}}, \glo{SurfGen\cite{surfgen}} & - & \glo{Dupont \etal\cite{dupont2022data}},  \glo{Diffusion-SDF\cite{chou2022diffusionsdf}}, \glo{3D-LDM\cite{nam20223d}}, \glo{Rodin\cite{wang2022rodin}}, \glo{Shue \etal\cite{shue20223d}} \\
                 & Mesh              & \glo{Ben-Hamu \etal\cite{ben2018multi}} & \glo{TM-Net\cite{gao2021tm}}, \glo{SDM-NET\cite{gao2019sdm}} & \glo{MeshDiffusion\cite{liu2023meshdiffusion}}, \glo{SLIDE\cite{lyu2023controllable}} \\
2D   & Depth/Normal Maps & \glo{S$^2$-GAN\cite{wang2016generative}}, \glo{RGBDGAN\cite{noguchi2019rgbd}}, \glo{DepthGAN\cite{shi20223d}} & - & - \\
                 & Voxel Grids       & \glo{VoxGRAF\cite{schwarz2022voxgraf}}, \glo{HoloGAN\cite{nguyen2019hologan}}, \glo{VON\cite{zhu2018visual}},  \glo{NGP\cite{chen2021towards}},\glo{PLATONICGAN\cite{Henzler2019ICCV}}, \com{BlockGAN\cite{nguyen2020blockgan}} & - & - \\
                 & Neural Fields     & \glo{GRAF\cite{graf}}, \glo{$\pi$-GAN\cite{pigan}}, \glo{CAMPARI\cite{campari}}, \glo{CIPS-3D\cite{cips3d}}, \glo{GOF\cite{generativeoccupancy}}, \glo{StyleNeRF\cite{stylenerf}}, \glo{StyleSDF\cite{stylesdf}}, \glo{GRAM\cite{gram}}, \glo{MVCGAN\cite{zhang2022multiview}}, \glo{GeoD\cite{geod}}, \glo{GRAM-HD\cite{xiang2022gramhd}}, \glo{EpiGRAF\cite{epigraf}}, \glo{PoF3D\cite{shi2023pof3d}}, \glo{GSN\cite{GSN}}, \glo{Pix2NeRF\cite{cai2022pix2nerf}}, \glo{Disentangled3D\cite{tewari2022disentangled3d}}, \glo{3D-GIF\cite{3DGIF}}, \glo{NeRF-VAE\cite{nerfvae}}, \glo{LOLNeRF\cite{Rebain2022CVPR}}, \com{DisCoScene\cite{xu2022discoscene}}, \com{GIRAFFE\cite{giraffe}}, \com{GIRAFFE HD\cite{xue2022giraffehd}}, \gloreli{ShadeGAN\cite{shadegan}}, \gloreli{Volux-GAN\cite{tan2022volux}}, \glohuman{3D-SGAN\cite{semanticguidedhuman}}, \gloseman{FENeRF\cite{fenerf}}, \gloseman{IDE-3D\cite{sun2022ide}}& - & -\\
                 & Hybrid           & \glo{EG3D\cite{eg3d}}, \glo{VolumeGAN\cite{xu2021volumegan}}, \glo{Next3D\cite{sun2022next3d}} & - & \glo{GAUDI\cite{bautista2022gaudi}}, \glo{RenderDiffusion\cite{anciukevivcius2022renderdiffusion}} \\
    \end{tblr}
    \vspace{-5pt}
\end{table*}

%% file: sections/05_image_generation.tex
\begin{figure*}[t]
    \centering
    \includegraphics[width=1\linewidth]{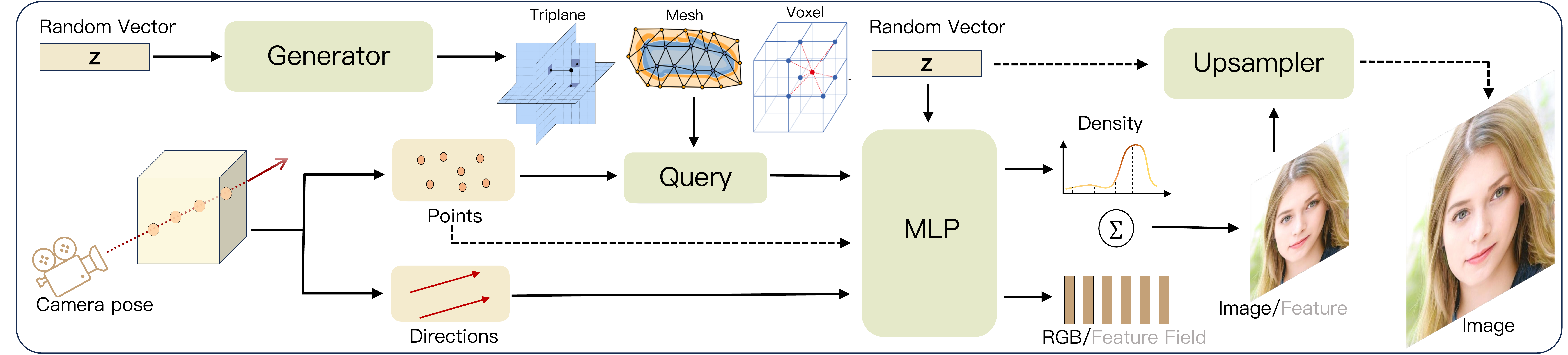}
    \caption{\textbf{The general pipeline of 3D-aware GAN.}
    The 3D-aware GAN framework generates 3D representations including Tri-plane~\cite{eg3d, epigraf}, Voxel~\cite{xu2021volumegan, schwarz2022voxgraf}, and Mesh~\cite{Gao2022NeurIPS}. These representations are then utilized to predict the color and density for volume rendering. The discriminator is omitted since it follows a similar approach as conventional 2D GANs.}
    \label{fig:3dgan_pipeline}
\end{figure*}

\section{Learning from 2D data}\label{sec:image_generation}
\begin{figure}
    \centering
    \includegraphics[width=1.0\linewidth]{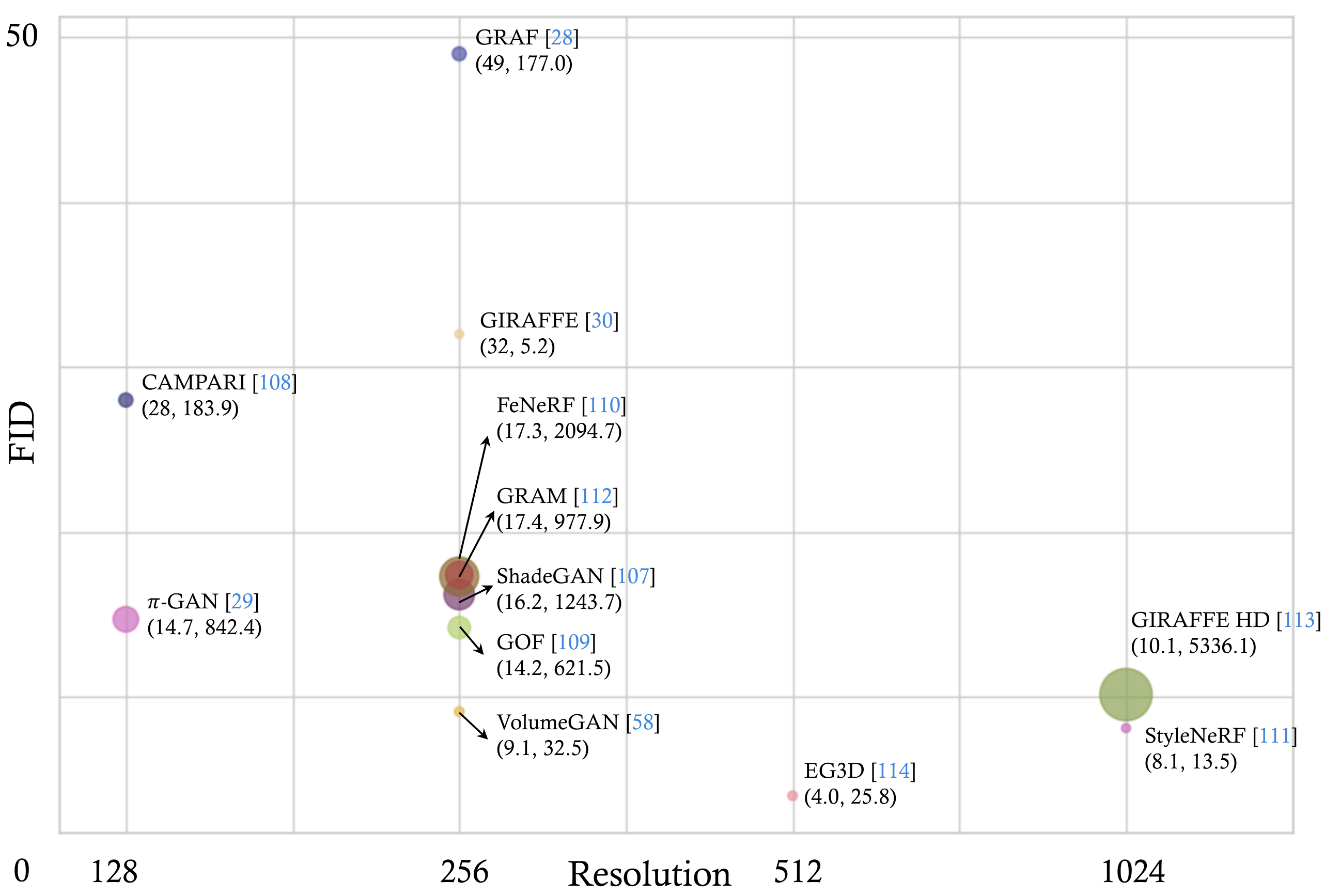}
    \caption{\textbf{FID \textit{v.s.} Resolution} of representative 3D synthesis methods trained on FFHQ~\cite{stylegan}. We annotate each method by their FID score and GigaFLOPS.}
    \label{fig:flops}
\end{figure}

The goal of learning 3D representation from 2D data is mostly to explicitly control the camera viewpoint when synthesizing images \cite{shen2020interpreting, tewari2020stylerig, deng2020disentangled, yang2021semantic, leimkuhler2021freestylegan}.
Although 2D GAN-based models deliver impressive results, finding a reasonable direction in the latent space is not easy and usually cannot support full control of the rendering viewpoint.
This survey focuses on works that explicitly generate 3D representations for 3D-aware image synthesis.
In contrast to 3D data-supervised methods that are directly trained with shapes, most 2D data-based generation methods are supervised by images through differentiable neural rendering because there are few high-quality and large-scale datasets of renderable 3D representations for training generative models.
Due to the lack of renderable 3D representations, auto-encoder architectures are rarely used in this task. Instead, most methods adopt generative adversarial models, which sample a latent vector from the latent space and decode it to the target representation, as shown in \cref{fig:3dgan_pipeline}.

Similar to 3D data-based generation, there are also several 3D representations commonly used in the task of 2D data-based 3D generation. These include depth/normal maps, voxel grids, and neural fields.
Point clouds and meshes are not well explored in generative image synthesis, partly because current differentiable neural rendering cannot provide effective gradient signals to optimize these two representations easily.
The key factors to consider when incorporating 3D representation for 2D image generation are quality and efficiency, where quality includes image realism and view consistency. Representative methods that learn from 2D data are summarized in \cref{tab:representation_classification}. Additionally, \cref{fig:flops} and \cref{tab:3dimage_generation} provide properties of representative methods.

\subsection{Depth/Normal Maps}

The depth and normal maps are easily accessible representations that partially reveal the geometry of 3D scenes or objects.
Since they only show the geometry from one side, they are usually referred to as 2.5D representations.
The depth and normal maps can be easily involved in image generation (\textit{i.e.}, processed by 2D convolutional neural networks rather than 3D architectures) as they share a similar data format to 2D images.
Most methods \cite{wang2016generative, noguchi2019rgbd, shi20223d} leverage GAN models to generate depth or normal maps for 3D-aware image synthesis.

S2-GAN \cite{wang2016generative} proposes to consider the underlying 3D geometry when generating images, where it refers to the geometry as "structure" and the image as "style". It develops the Structure-GAN, which maps a sampled latent vector to a surface normal map. The normal map is then fed into the Style-GAN with another sampled latent vector to generate the image. The Structure-GAN is implemented using a set of convolution and deconvolution layers, and the Style-GAN is designed based on the conditional GAN \cite{mirza2014conditional}. By disentangling the generation of the structure and style, S2-GAN makes the generative process more interpretable and enables the controllability of image synthesis given surface normal maps.
To explicitly control the viewpoint of rendered images, RGBD-GAN \cite{noguchi2019rgbd} passes camera parameters into the generator. It assumes that RGB and depth images can be generated simultaneously and learns from 2D image data only. To ensure the inter-view consistency of synthesized images, \cite{noguchi2019rgbd} generates two RGBD images from two different viewpoints and warps the RGB-D image from one view to another to compute the consistency loss. However, the model fails to capture this relationship when the contents in the image become more complex (\textit{e.g.}, bedrooms) and outputs a flat plane.
In contrast, Wang \etal \cite{wang2016generative} and Shi \etal \cite{shi20223d} argue that appearance should be rendered conditioned on the geometry, using normal maps and depth maps, respectively.
Similar to S2-GAN \cite{wang2016generative}, DepthGAN \cite{shi20223d} develops a dual-path generator for 3D-aware image synthesis. However, in contrast to S2-GAN, which explicitly conditions image generation on the synthesized geometry, DepthGAN extracts intermediate features from the geometry path and injects them into the image path.
DepthGAN \cite{shi20223d} also designs a switchable discriminator that not only classifies the RGB-D image but also regresses from the RGB image to the depth, which is then compared with the depth from the generator.
While depth or normal maps work well with 2D GAN models, which can efficiently synthesize high-resolution images, 2D CNN-based generators do not guarantee the inter-view consistency of generated images \cite{sitzmann2019srns}, even when supervised with the view consistency loss. Moreover, depth and normal maps are 2.5D representations that cannot fully capture the geometry of the scene.

\subsection{Voxel Grids}

There are generally two ways to synthesize images from voxel grids.
The first way is to use voxel grids only to represent 3D shapes, which are then used to render depth maps to guide image generation.
The second way is to embed the geometry and appearance of scenes with voxel grids.
Inspired by the graphics rendering pipeline, Zhu \etal \cite{zhu2018visual} and Chen \etal \cite{chen2021towards} decompose the image generation task into 3D shape generation and 2D image generation. They adopt voxel grids for shape generation, followed by a projection operator that renders voxel grids into the depth map under a viewpoint. The depth map is then used as guidance for RGB image generation. Similar to S2-GAN, Zhu \etal \cite{zhu2018visual} utilize a conditional GAN to synthesize the 2D image based on the rendered depth map. To enable illumination control, Chen \etal \cite{chen2021towards} generate different image components (\textit{e.g.}, diffuse albedo, specular, and roughness) based on the rendered depth and compose the final RGB output following the physical image formation process.
On the other hand, some methods \cite{nguyen2019hologan, nguyen2020blockgan, schwarz2022voxgraf} represent the underlying geometry and appearance in voxel grids simultaneously.
HoloGAN \cite{nguyen2019hologan} leverages a 3D CNN-based generator to produce a 3D feature volume from a latent vector. A projection unit then projects the voxels to 2D features, and the subsequent 2D decoder outputs the final RGB image. To control the viewpoint, HoloGAN applies 3D transformations to voxel grids, which ensures the underlying 3D geometric property.
Based on HoloGAN, BlockGAN \cite{nguyen2020blockgan} generates multiple feature volumes, each representing a 3D entity, and composes them into a feature volume for the whole 3D scene. This strategy enables it to render images containing multiple objects.
Instead of learning a 3D feature volume, PlatonicGAN directly predicts RGB and alpha volumes \cite{Henzler2019ICCV}, but is limited to a low resolution.
More recently, VoxGRAF \cite{schwarz2022voxgraf} attempts to synthesize RGB and alpha volumes and render images from volumes through volume rendering techniques. By removing the 2D CNN renderers, VoxGRAF promises to render view-consistent images. To achieve the rendering of high-resolution images, VoxGRAF adopts sparse voxel grids to represent scenes. It first generates a low-resolution dense volume and then progressively upsamples the resolution of the volume and prunes voxels in the empty space. 

\subsection{Neural Fields}

Image synthesis methods based on neural fields generally adopt MLP networks to implicitly represent the properties of each point in the 3D space, followed by a differentiable renderer to output an image under a specific viewpoint.%
The volume renderer~\cite{volumerendering} is the most commonly used renderer for 3D-aware image synthesis.
Most of the methods use GANs to supervise the learning of neural fields.
GRAF~\cite{graf} firstly introduces the concept of generative neural radiance fields. An MLP-based generator is conditioned on a shape noise and an appearance noise and predicts the density and the color of points along each ray. Then, a volume renderer gathers the information along all rays to synthesize a 2D image. Due to the slow rendering process, a patch-based discriminator is used to differentiate between the real and fake patches instead of the complete images.
$\pi$-GAN~\cite{pigan} uses a similar setting as GRAF, but it adopts SIREN~\cite{sitzmann2020siren} rather than ReLU MLPs for representation, which is more capable of modeling fine details. It utilizes progressive training with an image-based discriminator. Besides, instead of using two latent codes, $\pi$-GAN employs a StyleGAN-like mapping network to condition the SIREN on a latent code through FiLM conditioning~\cite{perez2018film,dumoulin2018featurefilm}.
Although the above methods improve the quality of 3D-aware image synthesis significantly, they still face several challenges. First, the good quality of rendered images does not imply a decent underlying shape and guarantees inter-view consistency. Second, due to the huge number of points queried along all rays and the complex rendering process, rendering an image from the model takes a lot of time. Consequently, it is hard to train a model on high-resolution images efficiently. Third, they all assume a prior distribution of camera poses on the dataset, which may not be accurate enough.

To mitigate the first issue, ShadeGAN~\cite{shadegan} adds a lighting constraint to the generator. It generates the albedo instead of the color for each point. A shading model operates on the generated albedo map, the normal map derived from densities, the sampled camera viewpoint, and the sampled lighting source to render an RGB image.
GOF~\cite{generativeoccupancy} analyzes the weight distribution on points along a ray. The cumulative rendering process cannot ensure a low-variance distribution on rays, leading to diffuse object surfaces. While occupancy-based neural fields can satisfy the requirement, they suffer from the optimization problem that gradients only exist on surfaces. Hence, GOF unifies the two representations by gradually shrinking the sampling region to a minimal neighboring region around the surface, resulting in neat surfaces and faster training.
GRAM~\cite{gram} also examines the point sampling strategy. It claims that the deficient point sampling strategy in prior work causes the inadequate expressiveness of the generator on fine details and that the noise caused by unstable Monte Carlo sampling leads to inefficient GAN training. GRAM proposes to regulate the learning of neural fields on 2D manifolds and only optimizes the ray-manifold intersection points.
MVCGAN~\cite{zhang2022multiview} attempts to alleviate the consistency problem by adding a multi-view constraint on the generator. For the same latent code, two camera poses are sampled for rendering, and both the features and images are warped from one to the other to calculate the reprojection loss.
GeoD~\cite{geod} tackles the problem by making the discriminator 3D-aware. A geometry branch and a consistency branch are added to the discriminator to extract shapes and evaluate consistency from the generator's outputs and in turn, supervise the generator.

To generate 3D-aware images on larger resolutions efficiently, a variety of methods operate neural fields in a small resolution and leverage convolutional neural networks for upsampling the image/features to a higher resolution.
StyleNeRF~\cite{stylenerf} integrates the neural fields into a style-based generator. The low-resolution NeRF outputs features rather than colors which are further processed by convolutional layers.
StyleSDF~\cite{stylesdf} merges a signed distance function-based representation into neural fields. The output features from the neural fields are passed to a style-based 2D generator for high-resolution generation. Two discriminators operate on the neural field and the 2D generator separately.
CIPS-3D~\cite{cips3d} uses a shallow NeRF network to get the features and densities of all points. A deep implicit neural representation network then predicts the color of each pixel independently based on the feature vector of that pixel. Each network is accompanied by a discriminator.
Despite the high-resolution images of high fidelity achieved, introducing convolutional layers for 2D upsampling may harm the 3D consistency. Carefully designed architectures and regularizers are necessary to mitigate the problem.
StyleNeRF~\cite{stylenerf} uses a well-thought-out upsampler based on 1x1 convolutional kernels. Moreover, a NeRF path regularizer is designed to force the output of CNNs to be similar to the output of NeRF.
GRAM-HD~\cite{xiang2022gramhd} designs a super-resolution module that upsamples the manifold to a higher resolution. Images are rendered from upsampled manifolds.
EpiGRAF~\cite{epigraf} bypasses the use of convolutional neural networks and instead operates directly on the output of a high-resolution neural field. It achieves this by introducing a patch-based discriminator that is modulated by the patch location and scale parameters.

Regarding camera poses, most methods assume a prior distribution from the dataset. 
Several methods\cite{gram,eg3d,xiang2022gramhd,epigraf,schwarz2022voxgraf}, however, leverage ground-truth poses of the dataset for sampling and training, resulting in a significant improvement in the quality of both the images and geometry. This demonstrates the importance of having accurate camera distribution. CAMPARI~\cite{campari} attempts to learn the pose distribution automatically from datasets and leverages a camera generator to learn the distribution shift based on a prior distribution. However, the results are sensitive to the prior initialization.
PoF3D~\cite{shi2023pof3d}, in contrast, infers the camera poses from the latent code and frees the model from prior knowledge. Besides, a pose-aware discriminator is designed to facilitate pose learning on the generator side.

With the advent of generative neural fields, researchers also explore compositional generation. 
GSN~\cite{GSN} decomposes global neural radiance fields into several local neural radiance fields to model parts of the whole scene.
GIRAFFE~\cite{giraffe} and GIRAFFE-HD~\cite{xue2022giraffehd} represent scenes as compositional generative neural radiance fields, allowing the independent control over multiple foreground objects as well as the background.
DisCoScene~\cite{xu2022discoscene} also proposes compositional generation but introduces scene layout prior to better disentangle and model the scene. 
UrbanGIRAFFE~\cite{yang2023urbangiraffe} further leverages panoptic layout prior to additionally enabling control over large camera movement and background semantics.

Besides, editing and controllability in the generation process have sparked the interest of researchers.
A few works focus on semantic control in 3D generation.
\cite{semanticguidedhuman} generates a semantic map from generative radiance fields conditioned on a semantic code as well as the sampled human pose and camera pose. It then employs an encoder-decoder structure to synthesize a human image based on the semantic map and a sampled texture latent code. The content can be manipulated by interpolating codes and sampling different poses.
FENeRF~\cite{fenerf} and IDE-3D~\cite{sun2022ide} render semantic maps and color images from generative neural fields simultaneously. The semantic maps can be used to edit the 3D neural fields via GAN inversion techniques~\cite{image2stylegan,alae}.
Pix2NeRF~\cite{cai2022pix2nerf} employs generative models to translate a single input image to radiance fields, granting control in the 3D space (\textit{e.g.}, changing viewpoints). To accomplish this, an encoder is added to the GAN framework to convert the input to the latent space, thereby establishing a reconstruction objective for learning. 
D3D~\cite{tewari2022disentangled3d} decomposes the generation into two distinct components: shape and appearance. This decomposition allows for independent control over geometry and color, respectively.
3D-GIF~\cite{3DGIF} and VoLux-GAN~\cite{tan2022volux} seek control over lights.
3D-GIF explicitly predicts the albedo, normal, and specular coefficients for each point. To synthesize an image that takes lighting into account, it employs volume rendering and photometric image rendering techniques using sampled lighting sources. 
VoLux-GAN also predicts albedos, normals, diffuse components, and specular components with the neural fields but instead uses an HDR environmental map to model the lighting. This approach enables VoLux-GAN to achieve high dynamic range imaging (HDRI) relighting capabilities.

In addition to GAN-based methods, NeRF-VAE~\cite{nerfvae} leverages a variational autoencoder~\cite{vae} to model the distribution of radiance fields. The model takes a set of images and their corresponding camera poses from the same scene as input and generates a latent variable. The latent variable serves as input for the generator to synthesize the content. Next, a reconstructed image is obtained by volume rendering the synthesized content using the input camera poses. The reconstructed images are expected to be similar to the input images.
LOLNeRF~\cite{Rebain2022CVPR} employs an auto-decoder to learn a 3D generative model, where pixel-level supervision can be applied in contrast to image-level supervision used in GAN-based methods.

\subsection{Hybrid Representations}

Implicit representations can effectively optimize the entire 3D shape based on 2D multi-view observations through differentiable rendering. 
As a result, many studies~\cite{graf,pigan,giraffe,stylenerf,Gao2022NeurIPS} opt to integrate implicit representations into the generator to achieve 3D-aware image synthesis using 2D supervision.
Implicit representations implemented by MLPs are memory-efficient but tend to have a long evaluation time. 
On the other hand, explicit representations like voxel grids and depth work effectively with CNNs and are efficient for high-resolution image rendering. However, they suffer from memory overhead and view consistency issues.
By combining implicit and explicit representations, their complementary benefits can be harnessed for 3D-aware image synthesis, potentially enhancing image quality and rendering efficiency, which is in line with the results from the benchmark by Wang \etal~\cite{wang2023benchmarking}.

VolumeGAN~\cite{xu2021volumegan} proposes to represent objects using structural and textural representations.
To capture the structure, they first construct a feature volume and then utilize the features queried within this volume, along with the raw coordinates, to learn generative implicit feature fields.
Additionally, a texture network similar to StyleGAN2 is employed to provide texture information to the feature fields. 
By utilizing explicit voxel grids and implicit feature fields, VolumeGAN enhances both the evaluation speed and the quality of synthetic images. 
However, it is challenging to improve the resolution of structural representation because a larger feature volume results in an increased computational burden.
Unlike VolumeGAN, EG3D~\cite{eg3d} introduces the use of tri-planes as explicit representations instead of voxels. 
EG3D adopts a 2D generator to generate the tri-planes, offering a computationally efficient approach to scale up to high resolutions. 
Besides, EG3D renders a low-resolution feature map that is upsampled to the target resolution using a neural renderer, i.e., a 2D CNN, further reducing the computational burden of the volume rendering.
Moreover, EG3D incorporates additional techniques, such as generator pose conditioning and dual discrimination, to address view inconsistencies.
Another approach, Next3D~\cite{sun2022next3d}, proposes the combination of explicit mesh-guided control and implicit volumetric representation, aiming to leverage the benefits of both controllability and quality.

Recently, a few works have started exploring diffusion models for 3D generation.
GAUDI~\cite{bautista2022gaudi} utilizes a latent representation that can be decoded into triplane features and camera poses for rendering. They then train a diffusion model on this latent space.
RenderDiffusion~\cite{anciukevivcius2022renderdiffusion} replaces U-Net, a common structure used in diffusion models, with a tri-plane feature-based network for denoising. This allows for learning 3D generation solely from 2D supervision.

It has been shown that hybrid representations are effective in producing high-quality 3D-aware images and 3D shapes.
As a result, subsequent works such as AvatarGen~\cite{zhang2022avatargen} and GNARF~\cite{ Bergman2022ARXIV} also adopt tri-plane representations for human synthesis.
However, there is still room for improvement,  particularly in terms of training speed and strategy. %
For instance, VolumeGAN relies on progressive training, which presents challenges when transitioning to another framework.

%% file: sections/06_applications.tex
\begin{figure}[t]
  \centering
  \includegraphics[width=\linewidth]{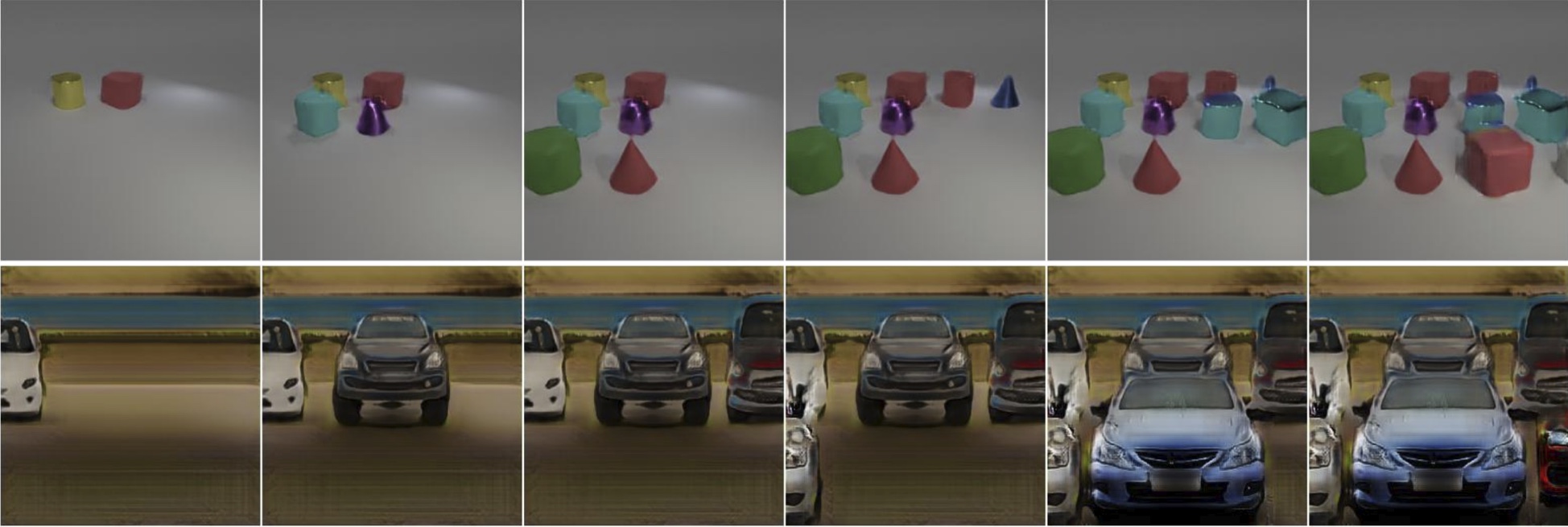}\\
  \vspace{-0.1cm}
  Layout editing 
  \includegraphics[width=\linewidth]{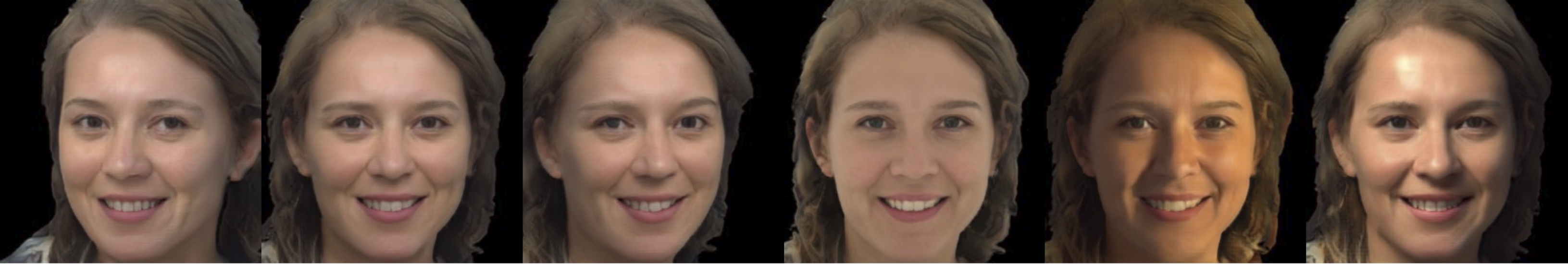}\\
  \vspace{-0.1cm}
  Relighting
  \includegraphics[width=\linewidth]{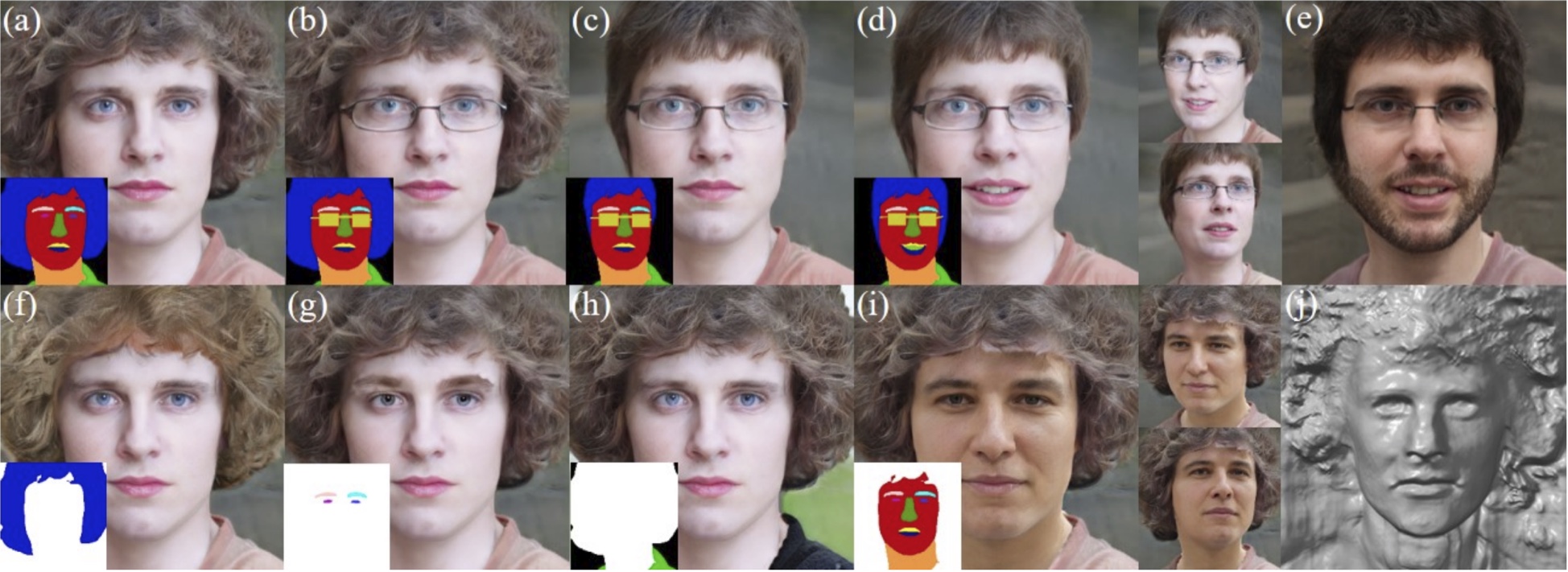}\\
  \vspace{-0.1cm}
  Semantic Editing
  \includegraphics[width=\linewidth]{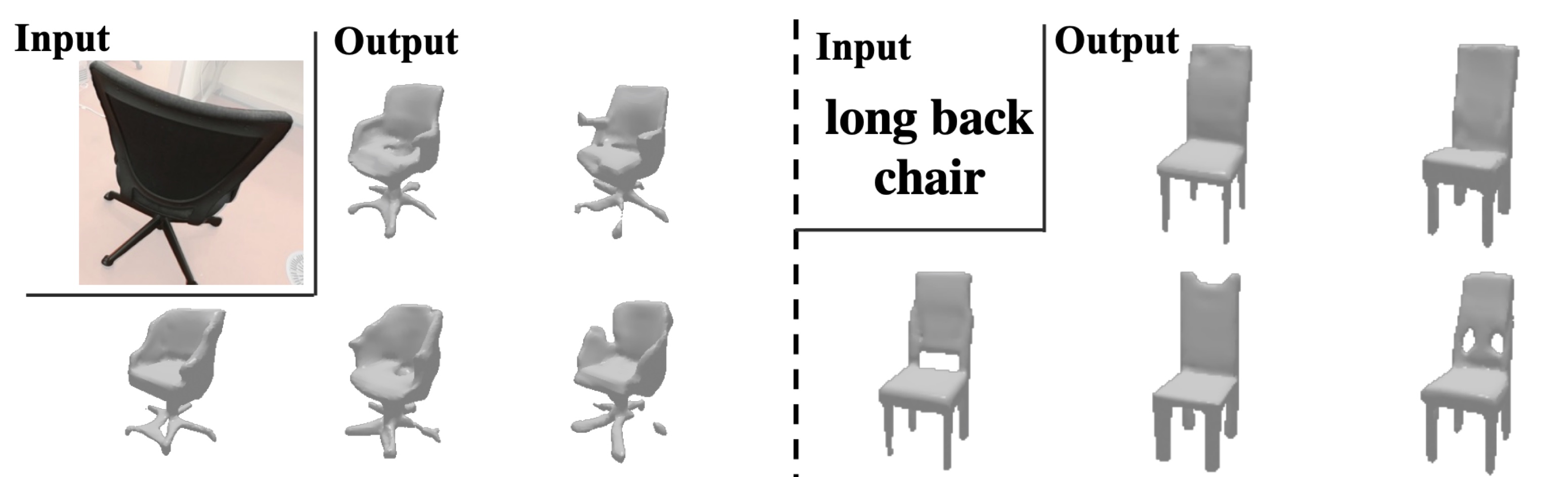}\\
  \vspace{-0.1cm}
  Reconstruction
\caption{\textbf{Applications}. 3D generative models have enabled various applications in editing, reconstruction, and representation learning. Images adapted from \cite{giraffe}\cite{tan2022volux}\cite{sun2022ide},\cite{mittal2022autosdf}. }
\label{fig:applications}
\end{figure}

\section{Applications}\label{sec:applications}

The emergence of 3D generative models has facilitated numerous promising applications, as shown in \cref{fig:applications}. In this section, we delve into the various applications of 3D generative models, specifically focusing on editing, reconstruction, and representation learning. Editing is further categorized into two distinct types: shape editing and 3D-aware image editing. This categorization is based on whether the resulting output exists in a 3D or 2D space.

\subsection{3D Shape Editing}\label{sec:shape_editing}

Several classical methods are available for 3D shape editing, including skeleton-based deformation and cage-based deformation. We refer to \cite{Yuan2021JCST} for a detailed survey of classical shape editing methods. Here, our focus lies on shape editing methods that apply to 3D   generative models.

\noindent\textbf{3D space editing: } Several works edit shapes relying on user manipulation in the 3D shape space. Liu et al.~\cite{Liu2017TDV} propose a method where users paint 3D shapes using voxel grids, which are then refined using a 3D GAN. Li et al.~\cite{li2020learning} allow users to edit a part of the voxel grid and automatically
deform and re-assemble all parts into a plausible shape. Alternatively, a few methods study shape editing controlled by more compact representations, such as sparse 3D points or bounding boxes~\cite{hao2020dualsdf,deng2021deformed,ibing20213d,Zheng2021CVPR}. While the aforementioned methods focus on editing rigid shapes, gDNA~\cite{gdna} enables reposing articulated human bodies controlled by human skeletons. 

\noindent\textbf{Latent code editing: } Another set of methods updates the latent space of the generative model without requiring user manipulation in the 3D space. Part-aware generative models that assign a latent code to each part inherently possess the ability of fine-grained part-level  control~\cite{gal2020mrgan,li2021sp}. Furthermore,
several methods explore shape editing guided by semantic labels or language descriptions~\cite{Jahan2021CGF,Liu2022CVPR,Fu2022ARXIV}.

\subsection{3D-Aware Image Editing}\label{sec:image_editing}
We refer to 3D-aware image editing as the technique that produces an edited 2D image as the outcome. We divide 3D-aware image editing methods into two categories: one that edits a set of physically meaningful 3D factors, while the other manipulates the latent code without explicit physical meaning.

\noindent\textbf{Physical factor editing: }
All 3D-aware image synthesis methods allow users to control camera poses, resulting in 2D images from various viewpoints~\cite{graf,pigan,gram,stylenerf,stylesdf,eg3d}. In addition to camera poses, several methods~\cite{nguyen2020blockgan,Liao2020CVPR,giraffe} enable the editing of object poses and the insertion of new objects by modeling the compositional nature of the scene. The study of re-lighting is also conducted through the disentangling of the albedo, normal, and specular coefficients~\cite{generativeoccupancy,3DGIF,tan2022volux}. More recently, several generative models for neural articulated radiance fields are proposed for human bodies~\cite{noguchi2022unsupervised,Bergman2022ARXIV,zhang2022avatargen}, enabling shape manipulation driven by human skeletons. 

\noindent\textbf{Latent code editing: }
The latent code captures all the other variations that are not represented by the physically meaningful 3D factors, e.g., object shape and appearance. One line of work involves editing the global attributes of the generated content. Several methods learn disentangled shape and appearance code for generative radiance fields, enabling independent interpolation of shape and appearance~\cite{graf,tewari2022disentangled3d}. CLIP-NeRF~\cite{Wang2022CVPR} further allows for controlling the shape and appearance of generative radiance fields using language or a guidance image. Another line of work explores more fine-grained local editing through semantic manipulation. FENeRF~\cite{fenerf} and IDE-3D~\cite{sun2022ide} allow users to edit the semantic masks and generate edited 3D-aware images accordingly.

\subsection{3D Reconstruction}

Two common approaches exist for utilizing 3D generative models in reconstruction, depending on whether test-time optimization is employed or not.

\noindent\textbf{Armotized inference:} 
Generative models equipped with an encoder, such as VAE, offer a natural means of reconstructing 3D shapes from a given input. In contrast to standard 3D reconstruction focusing on one-to-one mapping from the input to the target 3D shape, generative methods allow for sampling numerous possible shapes from the posterior distribution. This line of work is applicable to different types of input by simply switching the encoder. Common input choices are single-view image~\cite{Zhang2022ARXIV,mittal2022autosdf} and point cloud~\cite{Zhang2022ARXIV}.

\noindent\textbf{Model inversion:} 
Another line of methods leverages the inversion of generative models for reconstruction, typically requiring test-time optimization. By adopting an auto-decoder framework, a complete shape can be reconstructed from a partial observation via  optimizing the latent code to align with the partial observation\cite{deepsdf,Duggal2021ARXIV}.

While the aforementioned methods rely on 3D supervision, recent advancements demonstrate single-view image-based 3D reconstruction by inverting the 3D-aware image generative models using only 2D supervision~\cite{pigan,eg3d}. This approach aligns with the classical concept of Analysis by Synthesis~\cite{Vuille2006Vision}. Notably, by inverting 3D-aware image generation models, it becomes possible to recover camera poses, thereby enabling category-level object pose estimation~\cite{Chen2020ECCV,Guo2022ECCV,shi2023pof3d}.

\subsection{Representation Learning}

Another common application of generative models is to learn better representations for downstream tasks, e.g., classification or segmentation.
Notably, representation learning on point clouds has proven effective for tasks like model classification and semantic segmentation~\cite{achlioptas2018learning,Eckart2021CVPR,xie2021generative}. Similarly, generative models of voxel grids have been adopted for tasks such as model classification, shape recovery, and super-resolution~\cite{xie2020generative}.

%% file: sections/07_future_work.tex
\section{Future Work}\label{sec:future_work}

The development of 3D generative models has advanced rapidly, but there are still a lot of challenges to overcome before they can be used for downstream applications, such as gaming, simulation, and augmented/virtual reality. Here, we discuss current gaps in the literature and potential future directions of 3D generative models.

\noindent\textbf{Universality:} Most of the existing 3D generative models are trained on simple object-level datasets, e.g., ShapeNet for 3D shape generation and FFHQ for 3D-aware image synthesis. We believe that developing more universal 3D generative models is a fruitful direction for future research. Here, universality includes generating versatile objects (e.g., ImageNet or Microsoft CoCo), dynamic objects or scenes, and large-scale scenes. Instead of focusing on a single category, it is particularly interesting to learn a general 3D generative model for various categories similar to the 2D generative models, e.g.,  DALL-E 2 and Imagen~\cite{Ramesh2022ARXIV,Saharia2022ARXIV}.

\noindent\textbf{Controllability:} The controllability of 3D generative models lags behind that of 2D generative models. Ideally, the user should be able to control the 3D generation process via user-friendly input, including but not limited to language, sketch, and program. Moreover, we believe that the controllability of physical properties should be further improved, including lighting, material, and dynamics.

\noindent\textbf{Efficiency: } Many 3D generative models require 3-10 days of training on multiple high-end GPUs and are slow during inference. We believe that improving the training efficiency of 3D generative models is desirable to address the high impact of the models on the environment, and enhancing the inference efficiency is crucial for downstream applications.

\noindent\textbf{Training stability: } The training of 3D generative models, in particular 2D data-based generative models, is usually highly prone to mode collapse. One possible explanation is that the distribution of the physically meaningful factors, e.g., camera poses and rendering parameters, may not match that of the real images. Investigating the training stability of generative models is thus particularly important.

%% file: sections/08_conclusion.tex
\section{Conclusion}\label{sec:conclusion}

Deep 3D generative models, such as VAEs and GANs, aim to characterize the data distribution of observed 3D data.
A key challenge in the task of 3D generation is that there are many 3D representations for describing 3D instances, and each representation comes with its own advantage and disadvantage for generative modeling.
This paper presents a comprehensive review of 3D generation by discussing how different 3D representations combine with different generative models.
We first introduce the fundamentals of 3D generative models, including the formulation of generative models and 3D representations.
Then, we review how 3D representations are modeled and generated in the task of 3D generation learning from 2D and 3D data, respectively.
Afterward, we discuss the applications of 3D generative models, including shape editing, 3D-aware image manipulation, reconstruction, and representation learning.
Finally, we highlight the limitations of existing 3D generative models and propose several future directions.
We hope this review can help the readers to grasp a better understanding of the field of 3D generative models and inspire novel, innovative research in the future.